\documentclass{article}

\usepackage{amsmath,amssymb,amsfonts}
\usepackage{graphicx}
\usepackage{subcaption}
\usepackage{hyperref}
\usepackage{multirow}
\usepackage{makecell}
\usepackage{booktabs}
\usepackage{float}
\usepackage{xcolor}
\usepackage{changepage}
\usepackage{orcidlink}
\usepackage{arxiv}

\hypersetup{
	colorlinks=true,
	linkcolor=blue,
	citecolor=blue,
	urlcolor=blue
}

\title{Robust Transformer-Based One-Step Stock Index Forecasting via Shifted Data Augmentation}

\author{
	Tien Thanh Thach\,\orcidlink{0000-0001-7238-8778}\\
	Faculty of Mathematics and Statistics,
	Ton Duc Thang University,
	Ho Chi Minh City, Vietnam
}

\date{}

\renewcommand{\headeright}{}
\renewcommand{\undertitle}{}
\begin{document}
	
	\maketitle
	
	\begin{abstract}
		Transformers have shown remarkable success in sequence modeling, yet their direct application to financial time series remains challenging due to noisy signals, short-memory dynamics, and distributional shifts. This paper proposes a modified Transformer architecture for one-step stock index forecasting, combined with advanced learning-rate scheduling and a novel Shifted Data Augmentation (SDA) technique. We evaluate the proposed framework on two benchmark stock index datasets, VN30 and S\&P 500. Experimental results demonstrate that cosine annealing with warmup consistently improves forecasting accuracy over the generalized inverse-power scheduler. Furthermore, SDA substantially reduces forecasting errors and run-to-run variability while improving robustness to hyperparameter selection. The combination of cosine annealing scheduling and SDA achieved the best performance on both datasets, indicating that data augmentation can play a more important role than increasing model complexity in Transformer-based financial forecasting. These findings provide a practical and computationally efficient approach for robust stock index forecasting in noisy financial environments.
	\end{abstract}
	
	\textbf{Keywords:}
	Shifted Data Augmentation (SDA), Transformer, financial time series forecasting, learning-rate scheduling, stock index forecasting, VN30, S\&P 500.
	
	%=================================================================
	\section{Introduction}\label{Intro}

	Forecasting stock market indices is a long-standing challenge in financial research due to the inherent nonlinearity, volatility, and noise present in financial time series. Accurate predictions of index movements are of great importance for portfolio management, risk assessment, and algorithmic trading. Traditional statistical models, such as ARIMA and GARCH, often struggle to capture the complex nonlinear dynamics and long-range temporal dependencies present in financial data \cite{sezer2019survey}. In recent years, deep learning approaches have demonstrated significant promise for financial time series modeling, with recurrent neural networks (RNNs) and long short-term memory (LSTM) architectures being widely applied \cite{buczynski2023review}. However, these models face limitations in capturing long-range dependencies and parallelizing computations. Although recent efforts have introduced enhancements such as attention mechanisms and hybrid architectures \cite{kasse2025enhancing, hartanto2026attention, akkas2026attention}, these challenges remain. This motivates the exploration of Transformer-based approaches, which offer greater parallelization and improved modeling of global dependencies.
	
	The Transformer architecture, originally introduced for natural language processing (NLP) tasks \cite{vaswani2017attention}, has emerged as a powerful alternative for sequential modeling. By relying solely on attention mechanisms, Transformers can effectively capture global dependencies in sequences while enabling efficient parallelization. This has led to their adoption in diverse domains beyond NLP, including computer vision \cite{dosovitskiy2021image}, speech processing \cite{dong2018speech}, and time-series forecasting \cite{wen2023survey}. Recent studies have shown that Transformer-based models outperform traditional deep learning approaches in stock price prediction and financial time series modeling \cite{ieee2023lstmvstransformer, acm2024transformer}. Despite these advances, adapting Transformers to the specific characteristics of financial data remains an open research problem, requiring architectural modifications and optimization strategies tailored to this domain.
	
	In this paper, we focus specifically on one-step forecasting of stock market indices, i.e., predicting the next trading day's index value. Accurate one-step forecasts are important because they provide immediate guidance for short-term trading decisions, portfolio rebalancing, and risk management. Unlike multi-step forecasting, which may accumulate prediction errors over longer horizons \cite{bentaieb2012multistep, chevillon2007direct}, one-step forecasting minimizes error propagation and provides more direct short-term predictive signals. Even modest improvements in one-step forecast accuracy can be valuable in financial applications characterized by high volatility and sensitivity to market fluctuations. Moreover, reliable one-step forecasts can serve as building blocks for more complex strategies, including multi-horizon forecasting and scenario analysis, making one-step forecasting an important component of financial modeling.
	
	Guided by the challenges of forecasting noisy and nonlinear financial time series, our study is driven by the following research questions:
	
	\begin{enumerate}
		\item How can the Transformer architecture be adapted to effectively capture the nonlinear and noisy dynamics of financial time series for one-step stock index forecasting?
		
		\item What are the benefits of replacing the conventional ReLU activation with the Gaussian Error Linear Unit (GeLU) \cite{hendrycks2016gelu} in terms of predictive accuracy?
		
		\item How do different learning rate schedulers influence training stability and generalization in financial forecasting tasks?
		
		\item How can the Transformer be made more robust to hyperparameter sensitivity and distributional shifts inherent in financial markets?
	\end{enumerate}
	
	Unlike many previous studies that focus primarily on predictive accuracy, our investigation also emphasizes forecasting stability, reproducibility, and parameter efficiency, which are essential considerations for practical deployment in financial environments.
	To address these questions, we propose several innovations that extend and adapt the Transformer architecture for financial time series forecasting. The key contributions of our work are:
	
	\begin{itemize}
		\item \textbf{Architecture adaptation:} A modified Transformer tailored for forecasting stock market index values, bridging the gap between NLP applications and financial time series.
		
		\item \textbf{Activation function enhancement:} Integration of the GeLU activation function, which provides smoother nonlinearities and improved predictive accuracy compared to ReLU.
		
		\item \textbf{Regularization refinement:} A tailored dropout strategy that applies dropout to sublayer outputs before residual connections and layer normalization, while omitting dropout at the embedding stage. This modification improves training stability and forecasting accuracy in noisy financial data.
		
		\item \textbf{Advanced optimization strategies:} A generalized learning-rate scheduling framework that extends Vaswani et al.~\cite{vaswani2017attention} by introducing a tunable decay exponent $k$, alongside cosine annealing with warmup. These strategies provide greater flexibility in controlling optimization dynamics and improve training stability and forecasting accuracy.
		
		\item \textbf{Shifted Data Augmentation (SDA):} A novel augmentation strategy designed to improve robustness under distributional shifts in financial time series by exposing the model to shifted value ranges while preserving temporal structure.
		
		\item \textbf{Improved stability and reproducibility:} Extensive experiments across multiple independent runs demonstrate that SDA not only improves forecasting accuracy but also substantially reduces run-to-run variability, leading to highly stable and reproducible forecasting performance.
	\end{itemize}
	
	In addition to architectural and optimization innovations, SDA plays a central role in our approach. By replicating the dataset with constant offsets applied to the original index values while preserving temporal dependencies, SDA exposes the model to a wider range of scenarios and mitigates distributional mismatch between training and testing data. Experiments on the VN30 and S\&P 500 indices show that SDA substantially improves forecasting accuracy while dramatically reducing run-to-run variability. Furthermore, the resulting models exhibit reduced sensitivity to architectural hyperparameters, allowing compact Transformer configurations to achieve performance comparable to substantially larger networks. These findings suggest that combining Transformer adaptations, advanced optimization strategies, and SDA provides a practical and robust framework for real-world financial time-series forecasting.
	
	The remainder of this paper is organized as follows. Section~\ref{sec:background} reviews traditional forecasting approaches, recent deep-learning methods for financial forecasting, Transformer-based time-series models, and data augmentation techniques, and presents the motivation for the proposed approach. Section~\ref{sec:method} introduces the modified Transformer architecture, learning-rate scheduling strategies, and SDA. Section~\ref{sec:experimental} describes the datasets, data preparation and preprocessing procedures, model hyperparameters, and evaluation metrics. Section~\ref{sec:evaluation} presents the empirical evaluation and hyperparameter analysis. Section~\ref{sec:results} reports benchmark forecasting results on the VN30 and S\&P 500 datasets. Finally, Section~\ref{sec:concl} concludes the paper and discusses future research directions.
	
	%=================================================================
	\section{Background and Related Work}\label{sec:background}

	\subsection{Traditional Approaches}

	Financial time series forecasting has historically relied on statistical models such as autoregressive integrated moving average (ARIMA) and generalized autoregressive conditional heteroskedasticity (GARCH). These models are effective in capturing linear dependencies and volatility clustering, but they often fail to represent the nonlinear dynamics and long-range dependencies inherent in stock market indices \cite{sezer2019survey}. Extensions such as regime-switching models \cite{hamilton1989, kim1999} and hybrid econometric approaches \cite{diebold1996, stock2016} have also been proposed; however, their forecasting capability remains limited in highly volatile and nonstationary market environments \cite{akbal2024, lu2025}.
	
	\subsection{Deep Learning in Financial Forecasting}
	
	The emergence of deep learning introduced new possibilities for modeling financial time series. Recurrent neural networks (RNNs) and long short-term memory (LSTM) networks \cite{hochreiter1997long} have been widely adopted because of their ability to capture sequential dependencies \cite{buczynski2023review}. LSTMs, in particular, alleviate the vanishing gradient problem through gated memory mechanisms and have demonstrated improved one-step forecasting accuracy compared with traditional statistical models such as ARIMA in financial applications \cite{siami2018forecasting, siami2019comparative}.
	
	Variants such as gated recurrent units (GRUs) \cite{cho2014learning} and hybrid CNN-LSTM architectures have also been explored to capture both temporal and local features in financial data \cite{fischer2018deep, lu2020cnn}. More recently, encoder--decoder architectures integrated with attention mechanisms have demonstrated superior forecasting accuracy and stability compared with conventional LSTM and GRU models in stock market prediction tasks \cite{thach2025forecasting}.
	Despite these advances, RNN-based approaches remain limited by their sequential computation, which restricts parallelization and makes modeling very long sequences computationally inefficient \cite{vaswani2017attention}. Furthermore, when applied to noisy financial datasets, these models often exhibit sensitivity to hyperparameter selection and unstable training behavior, negatively affecting convergence and generalization.
	
	\subsection{Transformers for Financial Time Series}
	
	The Transformer architecture, introduced by Vaswani et al. \cite{vaswani2017attention}, revolutionized sequence modeling by replacing recurrence with self-attention mechanisms. Its ability to capture global dependencies while enabling efficient parallelization has led to widespread adoption in NLP and other domains. More recently, Transformers have been applied to time series forecasting, including financial applications, where they have demonstrated advantages over LSTMs in modeling long-range dependencies and improving forecasting accuracy \cite{ieee2023lstmvstransformer}. Several studies have proposed Transformer variants tailored to financial forecasting \cite{ieee2023lstmvstransformer, acm2024transformer, wang2022stock, bui2025time2vec, springer2024dualattention, psr2023phien, gezici2024, alridhawi2026, wang2023, friday2025}. 
	
	Existing approaches primarily emphasize architectural modifications and feature engineering. In contrast, our work focuses on optimization stability, reproducibility, and robustness to distributional shifts through advanced learning-rate scheduling and Shifted Data Augmentation (SDA).
	Furthermore, by evaluating the proposed framework on both the VN30 and the S\&P 500 indices, our study considers two contrasting financial environments: an emerging market that remains relatively underexplored in the literature and a mature market that serves as a global benchmark. This dual-market evaluation highlights the generalizability and practical relevance of the proposed approach across diverse financial conditions.
	
	\subsection{Data Augmentation in Time Series}
	
	Data augmentation has been widely studied in domains such as computer vision \cite{krizhevsky2012imagenet, perez2017augmentation, shorten2019augmentation} and natural language processing \cite{shorten2021augmentation, zhou2024augmentation}, but its application to time series forecasting remains comparatively limited. Existing augmentation strategies include noise injection, scaling, time warping, permutation, and generative approaches such as GANs and VAEs \cite{iglesias2023survey}. While these methods increase data diversity, they may distort temporal dependencies or introduce unrealistic patterns into financial sequences.
	
	To the best of our knowledge, constant-offset augmentation, where training sequences are replicated with additive shifts while preserving temporal structure, has not been explicitly explored in prior financial forecasting literature. Motivated by the nonstationary nature of financial markets, where future observations may occur at value levels not present in the training data, we introduce Shifted Data Augmentation (SDA), a simple yet effective strategy that expands the distribution of observed values without altering the underlying temporal dynamics. Experimental results in Sections~\ref{sec:evaluation} and~\ref{sec:results} demonstrate that SDA consistently improves forecasting accuracy, substantially reduces run-to-run variability, and decreases sensitivity to architectural hyperparameters in one-step forecasting tasks.
	
	\subsection{Motivation for Our Work}
	
	Although prior studies demonstrate the potential of Transformers in financial forecasting, most existing approaches adopt the original architecture with limited adaptation to the specific characteristics of financial time series. Consequently, important challenges remain unresolved, including optimization instability, hyperparameter sensitivity, vulnerability to distributional shifts, and excessive model complexity.
	Practical forecasting systems require not only predictive accuracy but also stable and reproducible behavior across independent training runs. Moreover, compact models are often preferred in deployment settings because of computational and latency constraints. These considerations motivate the development of Transformer architectures that are both robust and parameter-efficient. 
	
	To address these limitations, we propose a modified Transformer architecture incorporating GeLU activation, tailored dropout regularization, learning-rate scheduling, and the proposed SDA technique. Together, these modifications improve optimization stability, forecasting robustness, and predictive accuracy, while reducing sensitivity to architectural hyperparameters and enabling compact Transformer configurations to achieve competitive forecasting performance in one-step stock index forecasting.

	%=================================================================
	\section{Model Architecture}\label{sec:method}
	
	We consider the task of one-step forecasting in a univariate stock index time series $\{x_0, x_1, \ldots, x_T\},$ where the objective is to predict the next value $x_{T+1}$ based on the most recent $L$ observations. To formulate this forecasting problem within a supervised learning framework, we construct input--output pairs of the form
	\begin{align}\label{input-output}
		\mathbf{X}_t = [x_{t-L+1}, x_{t-L+2}, \ldots, x_t], \quad y_t = x_{t+1}, \quad t=L-1,\ldots,T-1.
	\end{align}
	The goal is to learn a nonlinear mapping
	\begin{align}
		f \colon \mathbb{R}^{L} \rightarrow \mathbb{R},
	\end{align}
	parameterized by a modified Transformer architecture, such that the predicted value
	\begin{align}
		\hat{y}_t = f(\mathbf{X}_t)
	\end{align}
	closely approximates the ground-truth target $y_t$.
	
	\begin{figure}[htbp]
		\centering
		\begin{subfigure}{0.5\textwidth}
			\includegraphics[width=\linewidth]{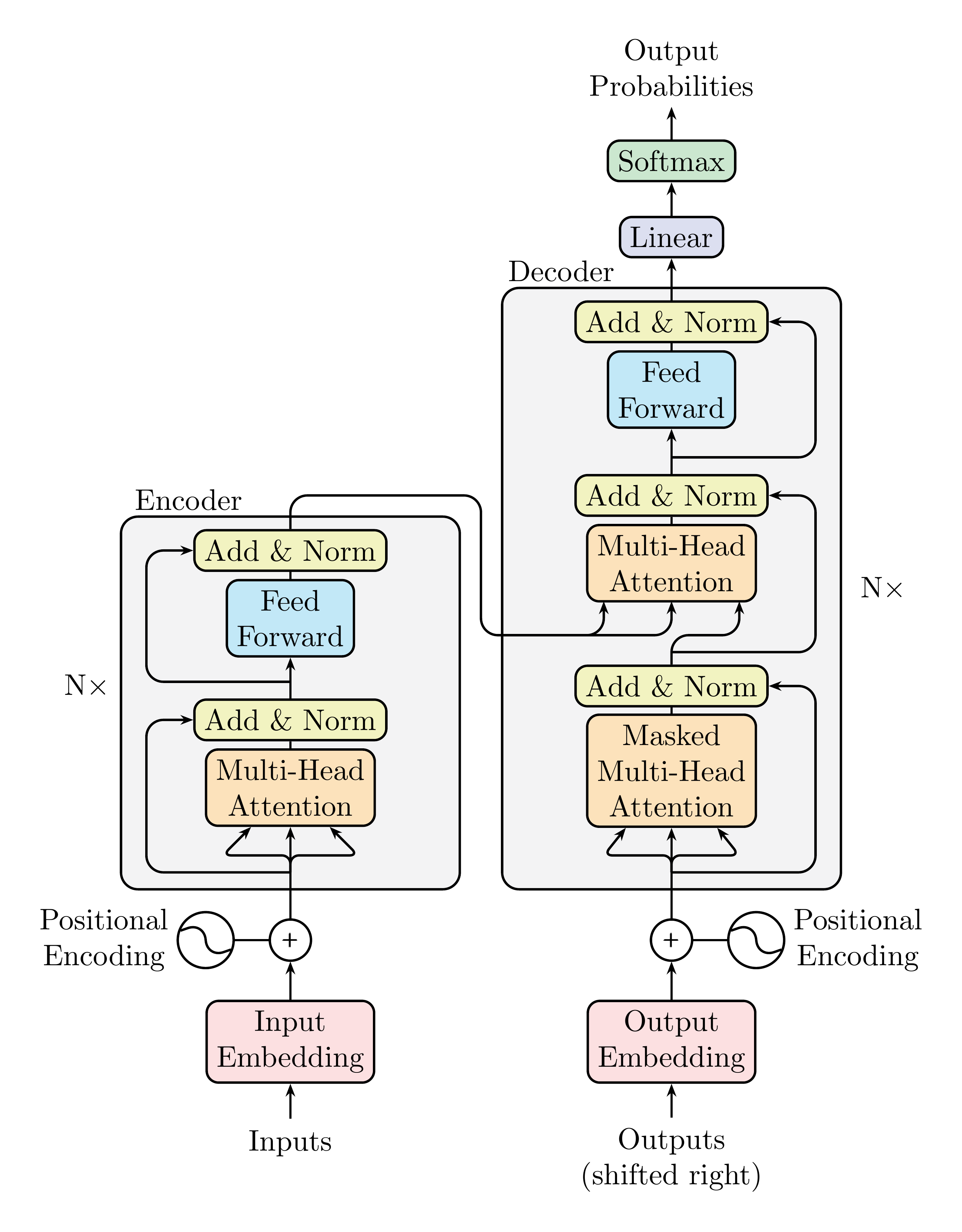}
			\caption{Original Transformer architecture.}
		\end{subfigure}\hfill
		\begin{subfigure}{0.5\textwidth}
			\vspace*{1.3cm}
			\includegraphics[width=\linewidth]{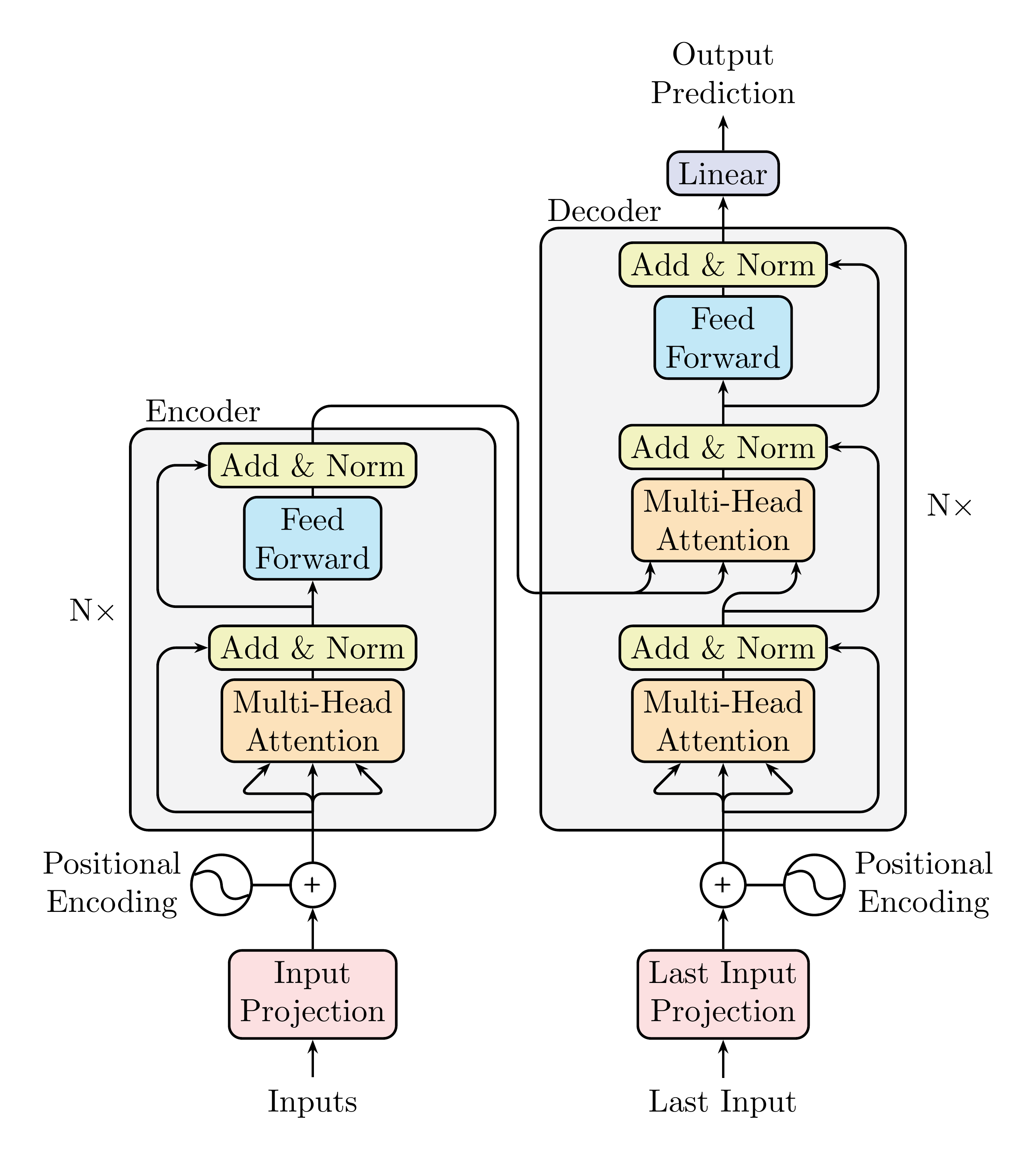}
			\caption{Modified Transformer for forecasting.}
		\end{subfigure}
		\caption{(a) Original Transformer architecture for NLP tasks and (b) proposed modified Transformer architecture for one-step stock index forecasting.}
		\label{fig:transf1}
	\end{figure}
	
	Figure~\ref{fig:transf1} compares the original Transformer architecture proposed by Vaswani et al.~\cite{vaswani2017attention}, originally designed for NLP tasks, with the modified Transformer architecture proposed in this work for one-step stock index forecasting. The original Transformer follows a sequence-to-sequence encoder--decoder framework that relies on token embeddings and softmax-based output generation. In contrast, our modified architecture is specifically adapted for continuous-valued financial time series.
	
	More precisely, scalar input observations are first projected into $d_{model}$-dimensional feature representations and combined with positional encodings to preserve temporal order information. The encoder then applies $N$ stacked layers consisting of multi-head self-attention and position-wise feed-forward networks. The decoder employs multi-head self-attention, encoder--decoder cross-attention, and feed-forward sublayers to model temporal dependencies and interactions across different time steps. Residual connections and layer normalization are incorporated throughout the network to improve optimization stability and convergence behavior.
	Unlike the original Transformer, which produces probability distributions over vocabulary tokens, the final output projection in the proposed architecture maps the decoder hidden states directly to scalar values, making the model suitable for one-step financial forecasting tasks.
	
	\subsection{Input Projection and Positional Encoding}
	
	Since Transformer architectures operate on sequences of vectors rather than scalar values, each input observation $x_t$ is first projected into a $d_{model}$-dimensional latent representation using a learned linear transformation:
	\begin{align}
		\mathbf{z}_t = x_t \mathbf{W}_{in} + \mathbf{b}_{in},
	\end{align}
	where $\mathbf{W}_{in} \in \mathbb{R}^{1 \times d_{\text{model}}}$ and $\mathbf{b}_{in} \in \mathbb{R}^{d_{\text{model}}}$ are learnable parameters. This operation plays a role analogous to token embeddings in the original Transformer architecture for NLP, but here it maps continuous-valued financial observations into the latent feature space of the model. Following Vaswani et al.~\cite{vaswani2017attention}, the projected vectors are scaled by $\sqrt{d_{\text{model}}}$.	This scaling helps maintain a consistent variance magnitude as the model dimension increases, thereby improving optimization stability and preventing excessively small gradients during attention computations.
	
	Unlike recurrent or convolutional architectures, the Transformer contains no inherent mechanism for encoding sequential order. To inject temporal information into the model, positional encodings are added to the projected inputs. Following Vaswani et al.~\cite{vaswani2017attention}, we employ deterministic sinusoidal positional encodings defined by
	\begin{align}
		PE(t,2i) &=	\sin\left( \frac{t}{10000^{2i/d_{\text{model}}}}\right), \\
		PE(t,2i+1) &= \cos\left(\frac{t}{10000^{2i/d_{\text{model}}}} \right),
	\end{align}
	where $t$ denotes the sequence position and $i$ indexes the embedding dimension. The use of sinusoidal functions with different frequencies enables the model to represent relative positional relationships through linear transformations of the encodings.
	The final embedded representation is obtained by combining the scaled projection with the positional encoding:
	\begin{align}
		\mathbf{e}_t = \sqrt{d_{\text{model}}} \cdot \mathbf{z}_t + PE(t).
	\end{align}
	This representation simultaneously captures feature information from the observed financial values and temporal order information from the positional encoding, allowing the Transformer to model sequential dependencies without relying on recurrence or convolution.
	
	\subsection{Encoder and Decoder Stacks}
	
	\textbf{Encoder:}
	The encoder consists of $N$ stacked layers, each containing two primary sublayers: a multi-head self-attention mechanism and a position-wise feed-forward neural network.
	The self-attention mechanism enables the model to capture temporal dependencies and interactions among observations within the historical input sequence. By attending to all positions simultaneously, the encoder can model both short-range and long-range relationships in financial time series.
	The resulting representations are then processed by a position-wise feed-forward network to enhance nonlinear feature extraction. Residual connections and layer normalization are applied after each sublayer to improve optimization stability and facilitate gradient propagation during training.
	
	\textbf{Decoder:}
	The decoder also consists of $N$ stacked layers. Each decoder layer contains three sublayers: a multi-head self-attention mechanism, an encoder-decoder multi-head attention mechanism, and a position-wise feed-forward neural network.
	Unlike the original Transformer designed for autoregressive sequence generation, the proposed framework focuses exclusively on one-step forecasting. Therefore, the decoder input is constructed simply from the last observed value in the input sequence, rather than from previously generated outputs. Consequently, causal masking is unnecessary, and the decoder self-attention mechanism becomes trivial because the input consists of a single value. The primary role of the decoder is therefore fulfilled by the encoder-decoder attention mechanism, which enables the decoder to attend to the temporal features extracted by the encoder from the historical input sequence. Residual connections and layer normalization are applied after each sublayer to stabilize training and improve convergence behavior.
	
	\textbf{Output Projection:}
	The final decoder representation is mapped back into the scalar prediction space using a linear output projection:
	\begin{align}
		\hat{y}_t =	\mathbf{h}_t \mathbf{W}_{out} + b_{out},
	\end{align}
	where $\mathbf{W}_{out} \in \mathbb{R}^{d_{\text{model}} \times 1}$ and $b_{out} \in \mathbb{R}$ are learnable parameters, and $\mathbf{h}_t$ denotes the final decoder hidden representation. The scalar output $\hat{y}_t$ represents the predicted next stock index value.
	
	\subsection{Scaled Dot-Product Attention}
	
	The core operation of the Transformer is the scaled dot-product attention mechanism, which enables the model to dynamically determine the importance of different time steps in the input sequence. In the context of financial forecasting, this allows the model to identify which historical observations are most relevant for predicting the next stock index value.
	The inputs to the attention mechanism are three matrices: queries $Q \in \mathbb{R}^{L \times d_k}$, keys $K \in \mathbb{R}^{L \times d_k}$, and values 
	$V \in \mathbb{R}^{L \times d_v}$. Each row of $Q$ is a query vector, each row of $K$ is a key vector, and each row of $V$ is a value vector. Here, $L$ denotes the sequence length, $d_k$ the dimension of queries and keys, and $d_v$ the dimension of values. The scaled dot-product attention is defined as
	\begin{align}
		\text{Attention}(Q,K,V)	= \text{Softmax}\left( \frac{QK^{T}}{\sqrt{d_k}} \right)V.
	\end{align}
	
\begin{figure}[htbp]
		\centering
		\includegraphics[width=0.7\linewidth]{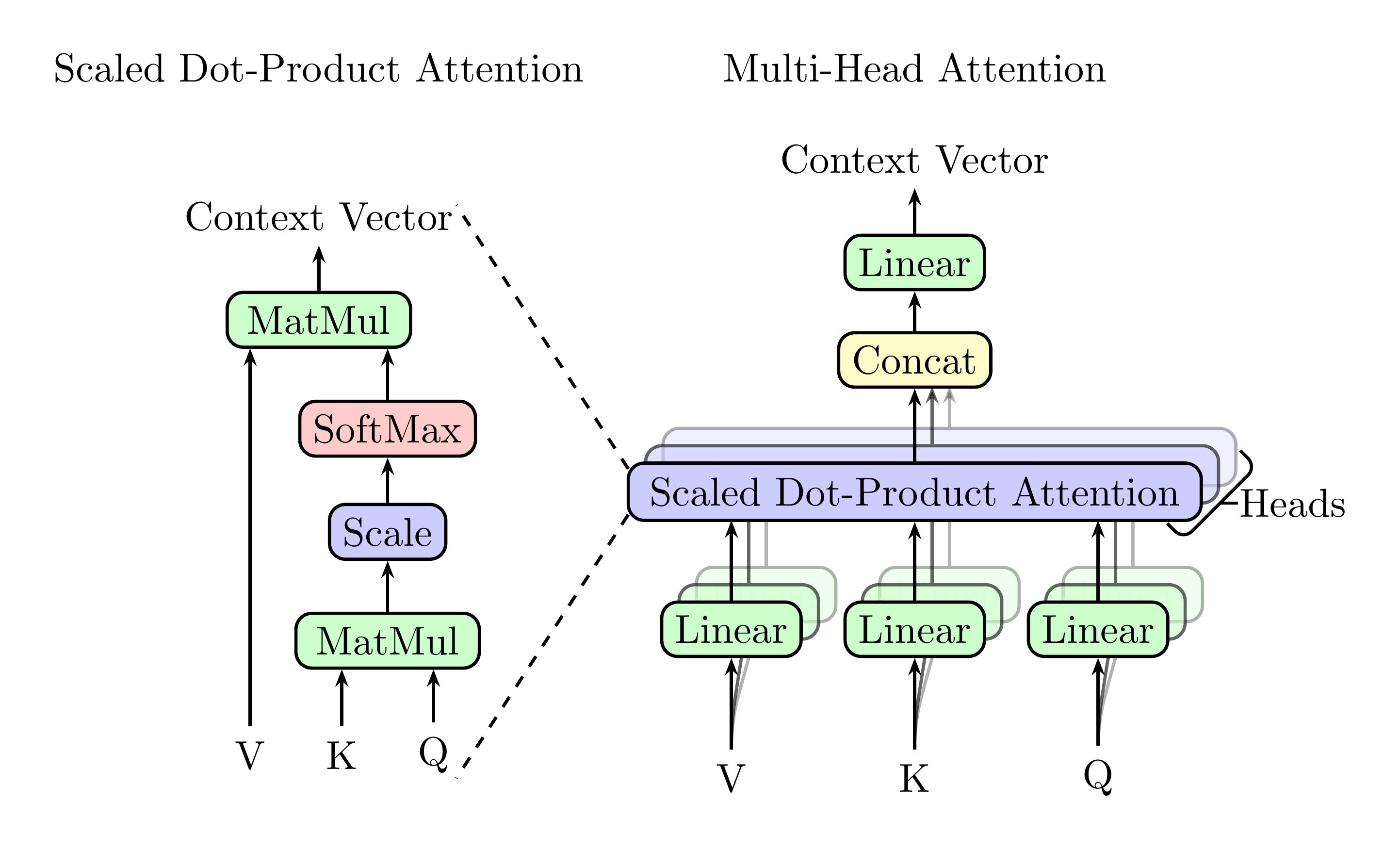}
		\caption{Illustration of the attention mechanisms used in the modified Transformer: (left) Scaled Dot-Product Attention and (right) Multi-Head Attention.}
		\label{fig:Attention}
	\end{figure}

	The matrix product $QK^{T}$ measures pairwise similarity between queries and keys, producing attention scores that quantify how strongly each time step attends to the others. Dividing by $\sqrt{d_k}$ prevents the dot products from becoming excessively large when the dimensionality increases, which would otherwise push the softmax function into regions with extremely small gradients and unstable optimization. After normalization through the softmax function, the resulting attention weights are used to compute a weighted combination of the value vectors. Consequently, the attention output can be interpreted as a context-aware representation of the input sequence, where more informative historical observations receive greater emphasis. Figure~\ref{fig:Attention} (left) illustrates the scaled dot-product attention mechanism, showing how queries, keys, and values interact to produce the context vector.
	
	\subsection{Multi-Head Attention}
	
\begin{figure}[htbp]
		\centering
		\includegraphics[width=0.8\linewidth]{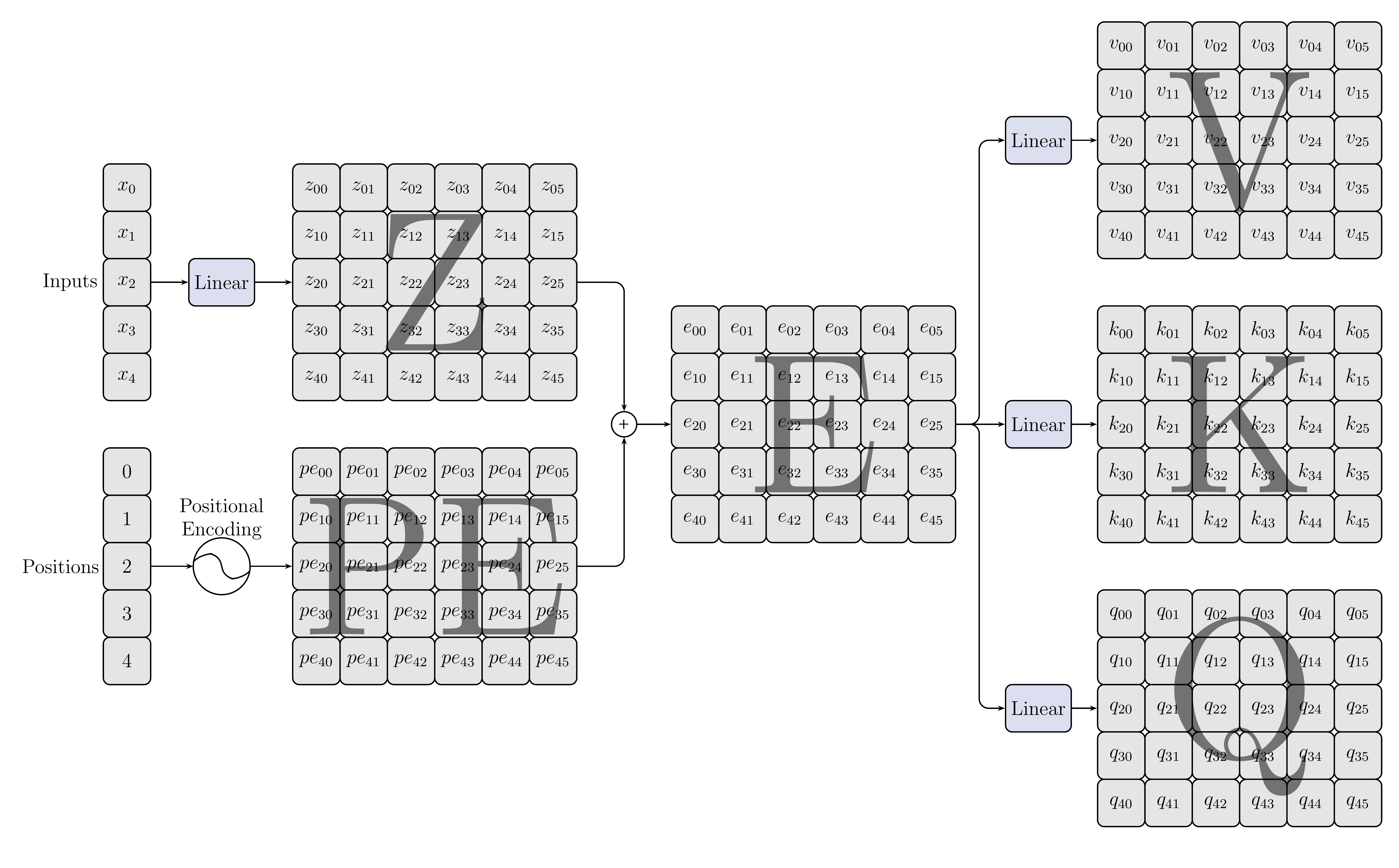}
		\caption{Illustration of projecting univariate time-series inputs into representation matrices $V$, $K$, and $Q$ with input length $L=5$ and model dimension $d_{model}=6$, including positional encodings.}
		\label{fig:projection}
	\end{figure}

	While the basic attention mechanism operates on a single representation space, the Transformer extends this idea through multi-head attention, which enables the model to capture different types of temporal relationships simultaneously. Specifically, the embedded sequence representations $\mathbf{e}_t$ are first projected into queries, keys, and values through learned linear transformations (Figure \ref{fig:projection}). These representations are then projected $h$ times into lower-dimensional subspaces of dimensions $d_k$, $d_k$, and $d_v$, respectively. The computation for the $i$-th attention head is given by
	\begin{align}
		\text{head}_i =	\text{Attention}(QW_i^Q,\; KW_i^K,\; VW_i^V),
	\end{align}
	where $W_i^Q \in \mathbb{R}^{d_{\text{model}} \times d_k}$, $W_i^K \in \mathbb{R}^{d_{\text{model}} \times d_k}$,	and	$W_i^V \in \mathbb{R}^{d_{\text{model}} \times d_v}$ are learned projection matrices associated with the $i$-th head.
	
	Each attention head independently learns relationships within its own representation subspace. In financial forecasting, different heads may focus on different temporal behaviors, such as short-term fluctuations, medium-term trends, or abrupt market changes. The outputs of all attention heads are concatenated and projected back into the model dimension:
	\begin{align}
		\text{MultiHead}(Q,K,V)	= \text{Concat}	( \text{head}_1,\ldots,	\text{head}_h)W^O,
	\end{align}
	where $W^O \in \mathbb{R}^{hd_v \times d_{\text{model}}}$ is the output projection matrix. This multi-head design enriches the representational capacity of the model by allowing several attention mechanisms to operate in parallel. Figure~\ref{fig:Attention} (right) illustrates how multiple attention heads are combined into a unified context representation.
	
	\subsection{Position-wise Feed-Forward Networks}
	
	Each encoder and decoder layer in the Transformer contains a position-wise feed-forward network (FFN), which applies the same nonlinear transformation independently to every position in the sequence. Unlike the attention mechanism, which models interactions across time steps, the FFN refines the representation at each position through nonlinear feature transformation. Formally, the FFN is defined as
	\begin{align}
		\text{FFN}(\mathbf{x})=	\text{GeLU}(\mathbf{x}\mathbf{W}_{1} + \mathbf{b}_{1}) \mathbf{W}_{2} + \mathbf{b}_{2},
	\end{align}
	where $\mathbf{x} \in \mathbb{R}^{d_{\text{model}}}$ is the input vector, $\mathbf{W}_{1} \in \mathbb{R}^{d_{\text{model}} \times d_{\text{ff}}}$, $\mathbf{W}_{2} \in \mathbb{R}^{d_{\text{ff}} \times d_{\text{model}}}$, and $\mathbf{b}_{1}, \mathbf{b}_{2}$ are learned bias vectors. The original Transformer architecture employs the Rectified Linear Unit (ReLU) activation function. In contrast, our modified architecture adopts the Gaussian Error Linear Unit (GeLU) \cite{hendrycks2016gelu}, defined as
	\begin{align}
		\text{GeLU}(x) = x\Phi(x),
	\end{align}
	where $\Phi(x)$ denotes the cumulative distribution function of the standard Gaussian distribution:
	\begin{align}
		\Phi(x)	= \frac{1}{\sqrt{2\pi}} \int_{-\infty}^{x} e^{-t^{2}/2} dt.
	\end{align}
	
	Unlike ReLU, which completely suppresses negative inputs, GeLU provides a smooth probabilistic gating mechanism that allows small negative values to contribute partially to the output. This smoother nonlinearity improves gradient propagation and reduces abrupt activation transitions during optimization.
	These properties are particularly beneficial for financial time series forecasting, where market signals are highly noisy, nonlinear, and sensitive to small fluctuations. Abrupt truncation of information may discard weak but informative patterns embedded in historical price movements. By preserving partial contributions from low-magnitude inputs, GeLU enables richer feature representations and more stable optimization dynamics. 
	GeLU has demonstrated strong empirical performance in large-scale Transformer-based architectures such as BERT \cite{devlin2018bert}. In our experiments, replacing ReLU with GeLU consistently improved forecasting accuracy and training stability across both the VN30 and S\&P 500 datasets.
	
	\subsection{Regularization}
	
	To reduce overfitting and improve generalization, dropout regularization was incorporated throughout the modified Transformer architecture. Following Vaswani et al.~\cite{vaswani2017attention}, dropout was applied to the outputs of each sublayer, including the multi-head attention and position-wise feed-forward networks, before the residual connections and layer normalization operations.
	However, our architecture differs from the original Transformer in one important aspect. In the original formulation, dropout is also applied after combining input projections with positional encodings. In contrast, we omit dropout at this stage.
	
	This modification was motivated by the characteristics of financial time series data. Unlike NLP tasks, where embeddings represent discrete semantic tokens, the projected inputs in our framework correspond directly to continuous numerical market observations. Applying dropout immediately after projection may therefore remove or distort fine-grained temporal information that is critical for short-term forecasting.
	
	Empirically, we observed that omitting dropout at the embedding stage resulted in more stable optimization, and consistently lower forecasting errors across both the VN30 and S\&P 500 datasets. At the same time, retaining dropout within the attention and feed-forward sublayers continued to provide sufficient regularization to prevent overfitting. This design preserves the integrity of low-level temporal representations while still benefiting from the regularization effects of dropout in deeper layers of the network.
	
	\subsection{Shifted Data Augmentation}
	
	To improve the robustness and generalization ability of Transformer models in one-step forecasting, we introduce a simple yet effective data augmentation technique termed \emph{Shifted Data Augmentation} (SDA). The key idea is to replicate the original training set by adding constant offsets to the index values, thereby exposing the model to shifted value ranges while preserving the temporal structure of the series. Formally, let $\mathbf{X} = \{x_t\}_{t=0}^{T}$ denote the original training sequence. We construct augmented sequences by applying additive shifts:
	\begin{align}
		\mathbf{X}^{(c)} = \{x_t + c \mid t = 0, \dots, T\},
	\end{align}
	where $c \in \mathcal{C}$ is a constant offset. In our experiments, we define the set of offsets as $\mathcal{C} = \{750\}$ for VN30 and $\mathcal{C} = \{4000\}$ for S\&P 500. The augmented training set is then given by
	\begin{align}
		\mathbf{X}_{\text{aug}}	= \mathbf{X} \cup \bigcup_{c \in \mathcal{C}} \mathbf{X}^{(c)}.
	\end{align}
	Unlike augmentation strategies that distort temporal dependencies through noise injection, permutation, or time warping, SDA preserves the sequential structure of the original series while expanding the range of values observed during training. Consequently, the model becomes less sensitive to scale variations and distributional shifts that frequently occur in financial markets.
	
	The motivation behind SDA arises from the non-stationary nature of financial time series. In practical forecasting scenarios, future market values may move beyond the range observed in the historical training data due to long-term market trends, macroeconomic changes, or regime shifts. Models trained only on a limited historical range may therefore generalize poorly when future values deviate substantially from past observations. By exposing the Transformer to shifted versions of the training sequence, SDA improves robustness to such distributional shifts while maintaining the underlying temporal dynamics. In practice, the offset constants are selected to expand the range of values observed during training and expose the model to plausible unseen value levels. Empirically, SDA substantially improves forecasting accuracy, reduces run-to-run variability, and decreases sensitivity to hyperparameter configurations in Transformer-based one-step forecasting tasks.
	
	\subsection{Loss Function and Optimization}
	
	The proposed model is trained to minimize the mean squared error (MSE) loss:
	\begin{align}
		L =	\frac{1}{T-L+1} \sum_{t=L-1}^{T-1} \left(	\hat{y}_t - y_t	\right)^2,
	\end{align}
	where $\hat{y}_t$ denotes the predicted value and $y_t$ the ground-truth target defined in Section~\ref{sec:method}. The MSE loss is widely used in regression and forecasting tasks because it strongly penalizes large prediction errors and provides stable gradients during optimization.
	Model parameters were optimized using the Adam optimizer \cite{kingma2014adam} with the default hyperparameters $\beta_1 = 0.9$, $\beta_2 = 0.999$, and $\epsilon = 10^{-8}$. Adam combines momentum and adaptive learning-rate estimation, making it particularly suitable for training deep neural networks on noisy financial time series.
	
	\subsubsection{Generalized Inverse-Power Learning Rate Scheduling}
	
	To regulate the optimization process, we first adopted the learning-rate schedule proposed by Vaswani et al.~\cite{vaswani2017attention} and generalized its decay exponent into a tunable parameter $k$. The learning rate at training step $step$ is defined as
	\begin{align}
		lr = d_{\text{model}}^{-k} \cdot\min\left(step^{-k},\;	step \cdot warmup\_steps^{-(1+k)} \right).
	\end{align}
	Here, $d_{model}$ denotes the Transformer model dimension, $warmup\_steps$ controls the duration of the warmup stage, and $k$ determines the decay behavior of the learning rate.
	
	This schedule consists of two phases. During the warmup stage, the learning rate increases approximately linearly with the training step, preventing unstable updates during the early stage of optimization when model parameters remain poorly initialized. After the warmup period, the learning rate decreases according to an inverse-power law controlled by the exponent $k$.
	Notably, setting $k = 0.5$ recovers the original inverse square-root schedule proposed in the Transformer architecture \cite{vaswani2017attention}. Smaller values of $k$ produce a slower decay and therefore maintain larger learning rates for longer periods, whereas larger values accelerate the decay process. Consequently, the generalized formulation provides additional flexibility for adapting optimization dynamics to different datasets and model configurations.
	
	\subsubsection{Cosine Annealing with Warmup}
	
	Although the generalized inverse-power schedule provides flexible control over learning-rate decay, we additionally investigated cosine annealing with warmup, which demonstrated superior empirical performance in our experiments. Let $lr_{base}$ denote the maximum learning rate reached after warmup, $lr_{min}$ the minimum learning rate, and $total\_steps$ the total number of training iterations.
	
	\textbf{Warmup phase ($step \leq warmup\_steps$):}
	During the warmup stage, the learning rate increases linearly from zero to $lr_{base}$:
	\begin{align}
		lr = lr_{base} \cdot \frac{step}{warmup\_steps}.
	\end{align}
	This gradual increase stabilizes optimization during the initial training stage and reduces the risk of unstable parameter updates.
	
	\textbf{Cosine decay phase ($step > warmup\_steps$):}
	After the warmup stage, the learning rate decreases smoothly according to a cosine schedule:
	\begin{align}
		lr = lr_{min} +	\frac{1}{2} (lr_{base} - lr_{min}) \left( 1	+ \cos \left(\pi \cdot \frac{step - warmup\_steps}{total\_steps - warmup\_steps}\right)\right).
	\end{align}
	
	Cosine annealing was originally introduced in SGDR \cite{loshchilov2016sgdr} and has since become widely adopted in deep learning and Transformer-based optimization. Compared with inverse-power schedules, cosine annealing provides a smoother and more gradual reduction in the learning rate after the warmup stage, which often improves convergence stability and generalization performance.
	These properties are particularly beneficial for financial forecasting, where optimization landscapes are highly noisy due to volatile and nonstationary market dynamics. In our experiments, cosine annealing with warmup consistently achieved lower forecasting errors and stable performance across both the VN30 and S\&P 500 datasets.

	%=================================================================
	\section{Experimental Setup}\label{sec:experimental}
	
	We evaluated the proposed approach on the VN30 and S\&P 500 indices to assess the forecasting performance of Transformer-based models under different hyperparameter settings, activation functions, learning-rate schedules, and SDA configurations. This experimental design facilitates a systematic investigation of the effects of architectural choices, optimization strategies, and data augmentation on one-step-ahead forecasting performance in financial time series. The hardware and software environments used in the experiments are summarized in Table~\ref{config}.
	
	\begin{table*}[htbp]
		\centering
		\caption{Summary of hardware and software configurations used in the experiments.}
		\label{config}
		\small
		\begin{tabular}{ll}
			\toprule
			\textbf{Hardware} & \\
			\hline
			CPU & Intel(R) Core(TM) i7-14700KF \\
			GPU & NVIDIA GeForce RTX 5070 Ti (16 GB GDDR7)\\
			\hline
			\textbf{Software} & \\
			\hline
			Miniconda & Conda 25.11.1 \\
			Python & Python 3.10.19 \\
			PyTorch & PyTorch 2.11.0 \\
			CUDA & CUDA 13.1 \\
			Operating System & Microsoft Windows 11  \\
			Other Libraries & NumPy 2.2.6, Pandas 2.3.3, scikit-learn 1.7.2, Plotly 6.5.2 \\
			\bottomrule
		\end{tabular}
	\end{table*}
	
	\subsection{Datasets}
	
	We evaluated the proposed models on two representative stock market indices: the VN30 from Vietnam and the S\&P 500 from the United States. The VN30 tracks the 30 largest and most liquid companies listed on the Ho Chi Minh Stock Exchange (HOSE), whereas the S\&P 500 represents 500 leading companies across diverse sectors of the U.S. economy. These indices were selected to provide contrasting market environments, allowing us to assess the robustness and generalizability of the proposed methods across both emerging and developed markets.
	
\begin{figure}[htbp]
		\centering
		\includegraphics[width=\linewidth]{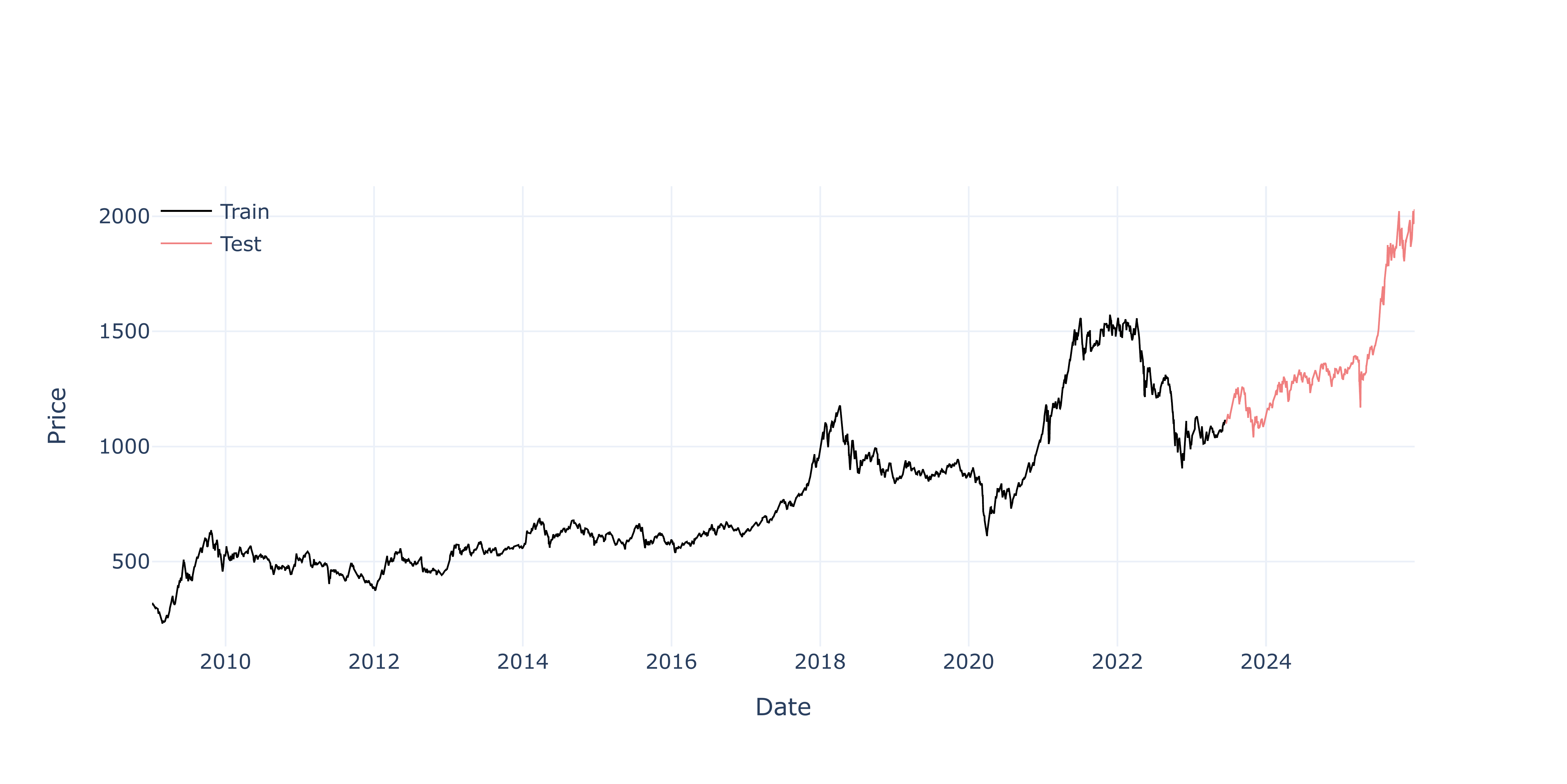}
		\caption{Time plot of VN30 index prices with train–test split.}
		\label{fig:vn30-indexplot}
	\end{figure}

\begin{figure}[htbp]
		\centering
		\includegraphics[width=\linewidth]{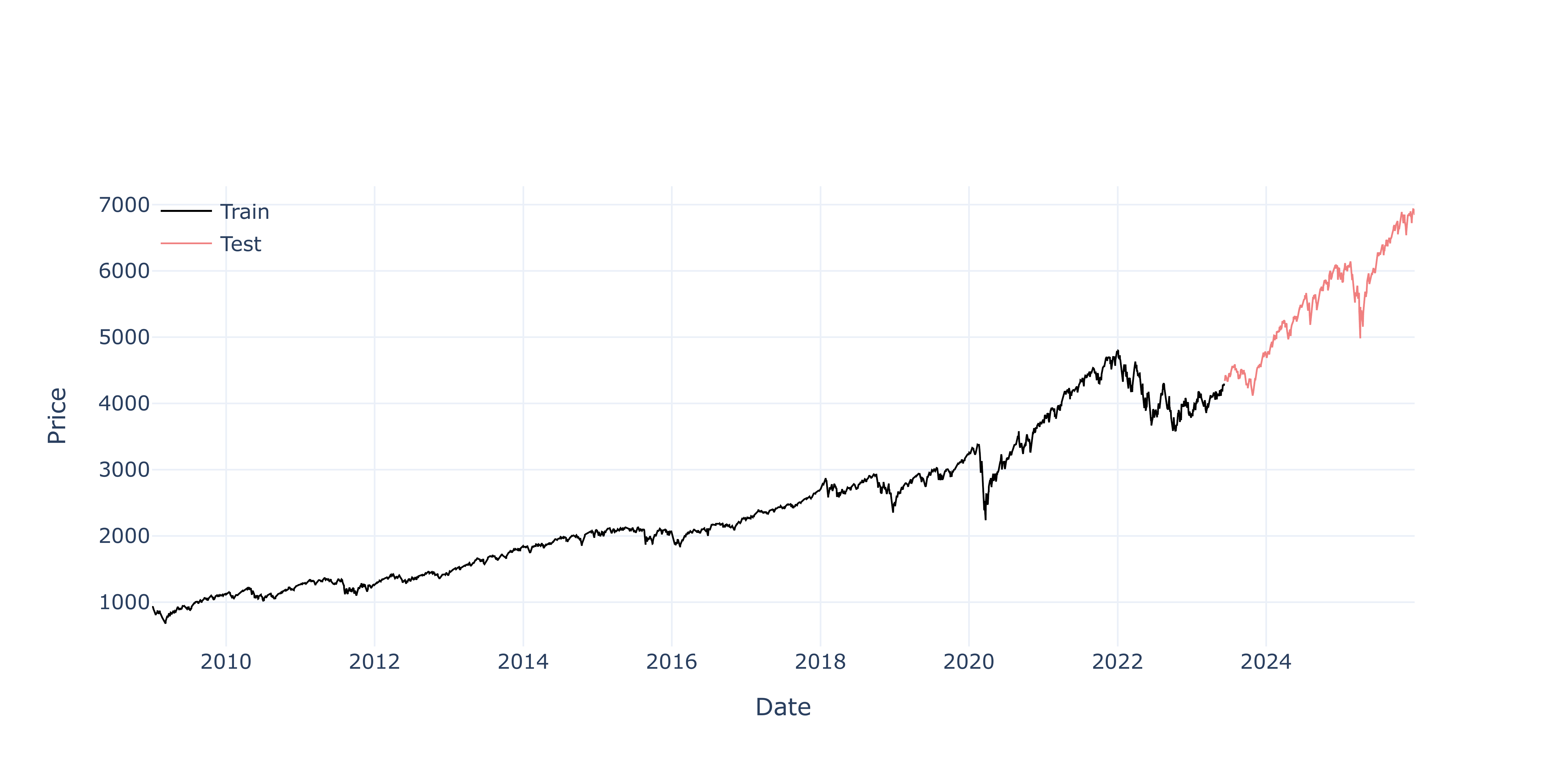}
		\caption{Time plot of S\&P 500 index prices with train–test split.}
		\label{fig:sp500-indexplot}
	\end{figure}

	Daily closing prices were collected for both indices from January 1, 2009 to December 31, 2025. This common period spans more than 16 years and includes multiple market regimes and economic cycles, providing a sufficiently long and diverse dataset for forecasting experiments.
	Figures~\ref{fig:vn30-indexplot} and~\ref{fig:sp500-indexplot} illustrate the historical trajectories of the VN30 and S\&P 500, respectively. Notably, the test sets contain values that exceed the range observed during training, creating a distributional shift that presents a challenging forecasting scenario and motivates the use of the proposed SDA method.
	
	\subsection{Data Preparation and Preprocessing}
	
	Each dataset was first divided into training and test sets, with the first 85\% of observations used for training and the remaining 15\% reserved for testing. To improve numerical stability and facilitate model training, all values were standardized using z-score normalization:
	\begin{align}
		x_t^\prime = \frac{x_t-\overline{x}}{s},
	\end{align}
	where $\overline{x}$ and $s$ denote the mean and standard deviation computed from the training set. The same statistics were applied to transform the test set, and the inverse transformation was used to recover predictions in the original scale.
	
\begin{figure}[htbp]
		\centering
		\includegraphics[width=0.8\linewidth]{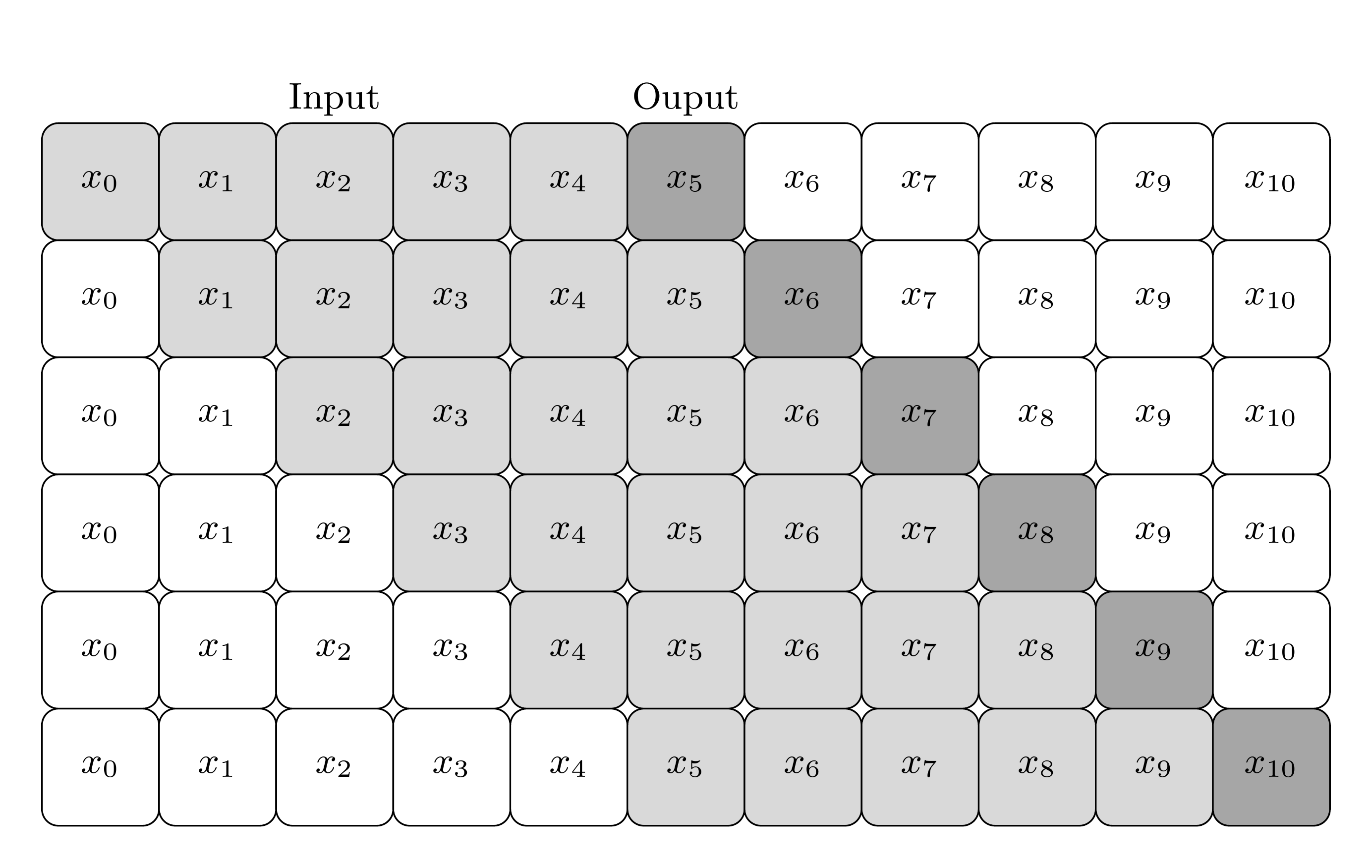}
		\caption{Illustration of the sliding window mechanism, showing overlapping input sequences and their corresponding outputs.}
		\label{fig:sliding_window}
	\end{figure}

	The standardized time series was then converted into supervised learning samples using a sliding-window approach. Given a sequence $\{x'_t\}$ and an input window length $L$, each sample consists of $L$ consecutive observations as input and the subsequent observation as the prediction target: $[x'_{t-L+1}, x'_{t-L+2}, \ldots, x'_t]\rightarrow x'_{t+1}$.	By sliding the window through the dataset, multiple overlapping input--output pairs were generated for training and evaluation. Figure~\ref{fig:sliding_window} illustrates the sample generation process.
	
	\subsection{Model Hyperparameters and Learning-Rate Schedules}
	
	The following hyperparameters were varied systematically during the experiments:
	\begin{itemize}
		\item $L$: input sequence length.
		\item $N$: number of encoder--decoder layers.
		\item $d_{\text{model}}$: model dimension.
		\item $d_{\text{ff}}$: feedforward network dimension.
		\item $h$: number of attention heads.
		\item $p_{\text{drop}}$: dropout rate.
		\item Activation function: ReLU or GeLU.
	\end{itemize}
	
	Two learning-rate schedules were evaluated:
	\begin{itemize}
		\item Generalized inverse-power schedule: the learning rate increases during warmup and subsequently decays proportionally to $(step)^{-k}$, where $k$ controls the decay rate.
		\item Cosine annealing with warmup: the learning rate increases linearly during warmup and then decays smoothly from $lr_{base}$ to $lr_{min}$ following a cosine schedule.
	\end{itemize}
	
	\subsection{Evaluation Metrics}
	
	To evaluate forecasting performance, we employed three widely used regression metrics: mean absolute error (MAE), root mean square error (RMSE), and mean absolute percentage error (MAPE). These metrics capture complementary aspects of prediction quality and are commonly used in financial time series forecasting.
	MAE measures the average magnitude of forecasting errors without considering their direction, providing an intuitive assessment of overall prediction accuracy. RMSE assigns larger penalties to large deviations due to the squared error term, making it particularly useful for detecting models that occasionally produce substantial forecasting errors. MAPE expresses prediction errors as percentages of the actual values, enabling scale-independent comparison across datasets with different numerical ranges.
	Formally, given actual values $y_i$ and predicted values $\hat{y}_i$ for $t = 1,\ldots,n$, the evaluation metrics are defined as follows:
	\begin{align}
		\text{MAE} &= \frac{1}{n} \sum_{i=1}^{n} \left|y_i - \hat{y}_i\right|,	\\
		\text{RMSE}	&= \sqrt{\frac{1}{n} \sum_{i=1}^{n}\left(y_i - \hat{y}_i	\right)^2},	\\
		\text{MAPE}	&= \frac{100}{n} \sum_{i=1}^{n} \left|\frac{y_i - \hat{y}_i	}{y_i} \right|.
	\end{align}
	Together, these metrics provide a comprehensive evaluation framework that captures average forecasting accuracy, sensitivity to large prediction errors, and relative error behavior across different market scales. This combination is particularly suitable for assessing the practical effectiveness of Transformer-based models in financial forecasting tasks.

	%=================================================================
	\section{Empirical Evaluation}\label{sec:evaluation}
	
	This section presents the experimental results of the proposed Transformer-based forecasting models on the VN30 and S\&P 500 datasets. To facilitate reproducibility, we report the key model hyperparameters, learning-rate scheduling strategies, and parameter counts. Model performance is evaluated using MAE, RMSE, and MAPE, with results averaged over multiple independent runs to reduce the effect of random initialization.
	
	\subsection{Baseline Transformer}
	
	The baseline Transformer serves as the reference model and corresponds to the modified Transformer architecture using the ReLU activation function, dropout regularization, and the inverse-power learning-rate scheduler described by Vaswani et al.~\cite{vaswani2017attention}. The scheduler was configured with $\text{warmup\_steps}=3000$, $\text{train\_steps}=15000$, and decay exponent $k=0.5$. 
	
	\begin{table}[htbp]
		\centering
		\caption{
			Evaluation metrics for the \textbf{VN30 dataset}, averaged over 10 independent runs of the baseline Transformer. The table reports forecasting performance (MAE, RMSE, and MAPE with standard deviations) under different architectural and optimization hyperparameter settings, together with approximate parameter counts.
		}
		\label{tab:VN30-table0}
		
		\scriptsize
		\setlength{\tabcolsep}{3pt}
		
		\resizebox{\textwidth}{!}{
			\begin{tabular}{c c c c c c c c c c | c c c | c}
				\toprule
				$L$ & $N$ & $d_{\text{model}}$ & $d_{\text{ff}}$ & $h$ & $d_k$ & $d_v$ & $p_{\text{drop}}$ & $k$ & Act. Fun. &
				MAE (SD) & RMSE (SD) & MAPE (SD) & \makecell{params \\ (approx)} \\
				\hline
				5 & 1 & 128 & 256 & 1 & 128 & 128 & 0.0 & 0.5 & ReLU
				& 47.38 (1.90) & 98.35 (4.84) & 2.71 (0.10) & 480K \\
				\hline
				&&&&&&& 0.1 &&& 51.29 (0.80) & 107.7 (1.33) & 2.92 (0.05) &\\
				&&&&&&& 0.2 &&& 49.34 (1.41) & 102.9 (2.12) & 2.83 (0.09) &\\
				&&&&&&& 0.3 &&& 44.43 (2.08) & 91.28 (5.61) & 2.58 (0.12) &\\
				&&&&&&& 0.4 &&& \textbf{42.24} (3.74) & \textbf{78.70} (6.33) & \textbf{2.55} (0.26) &\\
				&&&&&&& 0.5 &&& 56.94 (18.4) & 78.98 (15.0) & 3.77 (1.39) &\\
				\hline
				&&&& 2 & 64 & 64 &&&& 45.95 (2.58) & 95.06 (6.36) & 2.63 (0.14) &\\
				&&&& 4 & 32 & 32 &&&& 44.99 (2.39) & 93.10 (5.58) & 2.58 (0.13) &\\
				&&&& 8 & 16 & 16 &&&& 47.67 (3.49) & 98.40 (7.98) & 2.73 (0.19) &\\
				\hline
				& 2 &&&&&&&&& 54.23 (4.08) & 115.0 (9.98) & 3.07 (0.22) & 960K\\
				& 4 &&&&&&&&& 325.2 (258) & 390.6 (247) & 22.1 (18.5) & 1.9M\\
				\hline
				10 &&&&&&&&&& 46.67 (2.37) & 96.31 (5.62) & 2.67 (0.13) &\\
				20 &&&&&&&&&& 48.32 (2.91) & 99.48 (6.85) & 2.76 (0.15) &\\
				\hline
				&& 32 & 64 &&&&&&& 55.57 (4.23) & 118.8 (10.5) & 3.14 (0.22) & 31K\\
				&& 64 & 128 &&&&&&& 49.13 (3.00) & 102.6 (7.22) & 2.80 (0.16) & 121K\\
				&& 256 & 512 &&&&&&& 44.83 (3.21) & 91.21 (8.17) & 2.58 (0.16) & 1.9M\\
				&& 512 & 1024 &&&&&&& 42.35 (3.20) & 84.72 (7.87) & 2.46 (0.17) & 7.6M\\
				\bottomrule
			\end{tabular}
		}
	\end{table}
	
	\begin{table}[htbp]
		\centering
		\caption{
			Evaluation metrics for the \textbf{S\&P 500 dataset}, averaged over 10 independent runs of the baseline Transformer.}
		\label{tab:sp500-table0}
		\scriptsize
		\setlength{\tabcolsep}{3pt}
		
		\resizebox{\textwidth}{!}{
		\begin{tabular}{c c c c c c c c c c | c c c | c}
			\toprule
			$L$ & $N$ & $d_{\text{model}}$ & $d_{\text{ff}}$ & $h$ & $d_k$ & $d_v$ & $p_{\text{drop}}$ & $k$ & Act. Fun. &
			MAE (SD) & RMSE (SD) & MAPE (SD) & \makecell{params \\ (approx)} \\
			\hline
			5 & 1 & 128 & 256 & 1 & 128 & 128 & 0.0 & 0.5 & ReLU  
			& 492.5 (23.3) & 656.8 (28.7) & 8.04 (0.38) & 480K \\
			\hline
			&&&&&&& 0.1 &&& 554.5 (52.1) & 749.1 (65.5) & 9.02 (0.86) &\\
			&&&&&&& 0.2 &&& 532.2 (44.3) & 729.4 (57.4) & 8.64 (0.72) &\\
			&&&&&&& 0.3 &&& 514.3 (66.5) & 715.9 (84.6) & 8.34 (1.08) &\\
			&&&&&&& 0.4 &&& \textbf{422.5} (41.1) & \textbf{600.9} (56.2) & \textbf{6.87} (0.65) &\\
			&&&&&&& 0.5 &&& 425.9 (65.3) & 605.2 (83.3) & 6.93 (1.07) &\\
			\hline
			&&&& 2 & 64 & 64 &&&& 492.5 (22.3) & 659.6 (29.2) & 8.03 (0.36) &\\
			&&&& 4 & 32 & 32 &&&& 485.7 (20.7) & 650.0 (25.9) & 7.92 (0.34) &\\
			&&&& 8 & 16 & 16 &&&& 506.7 (35.0) & 675.3 (43.6) & 8.27 (0.58) &\\
			\hline
			& 2 &&&&&&&&& 590.2 (36.3) & 782.5 (46.2) & 9.64 (0.60) & 960K\\
			& 4 &&&&&&&&& 2577.9 (841) & 2759.8 (753) & 45.4 (15.7) & 1.9M\\
			\hline
			10 &&&&&&&&&& 479.0 (28.3) & 639.0 (36.3) & 7.81 (0.46) &\\
			20 &&&&&&&&&& 491.8 (27.7) & 650.2 (34.5) & 8.02 (0.46) &\\
			\hline
			&& 32 & 64 &&&&&&& 595.4 (89.9) & 810.6 (119) & 9.68 (1.47) & 31K\\
			&& 64 & 128 &&&&&&& 478.5 (56.3) & 644.6 (72.1) & 7.80 (0.92) & 121K\\
			&& 256 & 512 &&&&&&& 492.9 (65.2) & 652.7 (81.4) & 8.06 (1.08) & 1.9M\\
			&& 512 & 1024 &&&&&&& 489.6 (117) & 645.9 (148) & 8.01 (1.93) & 7.6M\\
			\bottomrule
		\end{tabular}
	}
	\end{table}
	
	Tables~\ref{tab:VN30-table0} and~\ref{tab:sp500-table0} summarize the forecasting performance, model hyperparameters, and parameter counts. Several observations can be drawn:
	\begin{itemize}	
		\item \textbf{Dropout rate ($p_{\text{drop}}$):}
		Moderate dropout substantially improved forecasting performance on both datasets, with the best results obtained at $p_{\text{drop}}=0.4$. Increasing the dropout rate further to $0.5$ led to larger variance and reduced stability, particularly for the VN30 dataset.
		
		\item \textbf{Number of attention heads ($h$):}
		Performance improved as the number of attention heads increased from 1 to 4, but deteriorated at $h=8$. These results suggest that a relatively small number of heads is sufficient for the one-step forecasting task considered in this study.
		
		\item \textbf{Model depth ($N$):}
		A single encoder--decoder layer consistently outperformed deeper configurations. Increasing depth resulted in higher prediction errors and substantially greater variance, indicating optimization difficulties and potential overfitting on the available training data.
		
		\item \textbf{Model dimension ($d_{\text{model}}$):}
		For VN30, forecasting accuracy generally improved as $d_{\text{model}}$ increased, with the best performance achieved at $d_{\text{model}}=512$. In contrast, the S\&P 500 obtained its best results with moderate dimensions ($d_{\text{model}}=64$--$128$), suggesting that larger representations do not necessarily translate into better forecasting performance.
		
		\item \textbf{Sequence length ($L$):}
		The shortest input window ($L=5$) consistently produced the best results on both datasets, while longer windows ($L=10$ and $20$) provided little or no improvement. This finding suggests that most predictive information is concentrated in recent observations.
		
		\item \textbf{Parameter efficiency:}
		Forecasting accuracy did not scale proportionally with model size. Although larger models generally improved performance on the VN30 dataset, the gains diminished as the parameter count increased from approximately 1.9M to 7.6M. For the S\&P 500 dataset, moderate-sized models ($d_{\text{model}}=64$--$128$) achieved the best overall results. This indicates that substantially larger Transformer models may offer limited additional predictive benefit relative to their computational cost in one-step financial forecasting.
	\end{itemize}
	
	\subsection{Modified Transformer with Generalized Inverse-Power Scheduling}
	
	The modified Transformer was trained using the generalized inverse-power learning-rate scheduler with $\text{warmup\_steps}=5000$, $\text{train\_steps}=15000$, and decay exponent $k=0.9$. 
		
	\begin{table}[htbp]
		\centering
		\caption{
			Evaluation metrics for the \textbf{VN30 dataset}, averaged over 10 independent runs of the modified Transformer with the generalized inverse-power scheduler.
		}
		\label{tab:VN30-table1}
		\scriptsize
		\setlength{\tabcolsep}{3pt}
		
		\resizebox{\textwidth}{!}{
		\begin{tabular}{c c c c c c c c c c | c c c | c}
			\toprule
			$L$ & $N$ & $d_{\text{model}}$ & $d_{\text{ff}}$ & $h$ & $d_k$ & $d_v$ & $p_{\text{drop}}$ & $k$ & Act. Fun. &
			MAE (SD) & RMSE (SD) & MAPE (SD) & \makecell{params \\ (approx)} \\
			\hline
			5 & 1 & 128 & 256 & 1 & 128 & 128 & 0.0 & 0.9 & GeLU  
			& 24.65 (1.40) & 44.90 (3.14) & 1.50 (0.07) & 480K \\
			\hline
			&&&&&&&& 0.5 && 44.47 (1.74) & 92.16 (4.33) & 2.55 (0.09) &\\
			&&&&&&&& 0.6 && 32.41 (1.22) & 63.20 (3.00) & 1.91 (0.06) &\\
			&&&&&&&& 0.7 && 29.17 (1.39) & 55.37 (3.29) & 1.74 (0.07) &\\
			&&&&&&&& 0.8 && 27.60 (1.00) & 51.71 (2.34) & 1.66 (0.05) &\\
			&&&&&&&& 0.8 & ReLU & 37.76 (0.69) & 75.25 (1.65) & 2.20 (0.04) &\\
			\hline
			&&&&&&& 0.1 &&& 33.36 (0.96) & 63.18 (1.99) & 1.99 (0.06) &\\
			&&&&&&& 0.2 &&& 34.83 (0.69) & 64.56 (2.27) & 2.09 (0.04) &\\
			&&&&&&& 0.3 &&& 35.51 (1.23) & 64.39 (2.17) & 2.14 (0.08) &\\
			\hline
			&&&& 2 & 64 & 64 &&&& 24.96 (1.99) & 45.58 (4.49) & 1.52 (0.11) &\\
			&&&& 4 & 32 & 32 &&&& 24.59 (2.51) & 44.78 (5.68) & 1.50 (0.13) &\\
			&&&& 8 & 16 & 16 &&&& 23.52 (1.42) & 42.35 (3.22) & 1.44 (0.08) &\\
			&&&& 16 & 8 & 8 &&&& \textbf{22.13} (1.66) & \textbf{39.24} (3.74) & \textbf{1.37} (0.09) &\\
			\hline
			& 2 &&&&&&&&& 23.82 (1.58) & 43.06 (3.56) & 1.46 (0.08) & 960K\\
			& 4 &&&&&&&&& 25.16 (1.34) & 46.24 (3.14) & 1.53 (0.07) & 1.9M\\
			\hline
			10 &&&&&&&&&& 25.38 (1.67) & 46.39 (3.75) & 1.54 (0.09) &\\
			20 &&&&&&&&&& 24.90 (1.05) & 45.00 (2.33) & 1.52 (0.06) &\\
			\hline
			&& 32 & 64 &&&&&&& 33.44 (2.40) & 65.64 (5.73) & 1.97 (0.13) & 31K\\
			&& 64 & 128 &&&&&&& 27.43 (1.71) & 51.39 (4.01) & 1.65 (0.09) & 121K\\
			&& 256 & 512 &&&&&&& 24.93 (1.41) & 45.42 (3.16) & 1.52 (0.08) & 1.9M\\
			&& 512 & 1024 &&&&&&& 23.72 (0.49) & 42.62 (1.11) & 1.46 (0.03) & 7.6M\\
			&& 1024 & 2048 &&&&&&& 26.41 (0.72) & 48.66 (1.60) & 1.60 (0.04) & 30M\\
			\bottomrule
		\end{tabular}
	}
	\end{table}
	
	\begin{table}[htbp]
		\centering
		\caption{
			Evaluation metrics for the \textbf{S\&P 500 dataset}, averaged over 10 independent runs of the modified Transformer with the generalized inverse-power scheduler.
		}
		\label{tab:sp500-table1}
		\scriptsize
		\setlength{\tabcolsep}{3pt}
		
		\resizebox{\textwidth}{!}{
		\begin{tabular}{c c c c c c c c c c | c c c | c}
			\toprule
			$L$ & $N$ & $d_{\text{model}}$ & $d_{\text{ff}}$ & $h$ & $d_k$ & $d_v$ & $p_{\text{drop}}$ & $k$ & Act. Fun. &
			MAE (SD) & RMSE (SD) & MAPE (SD) & \makecell{params \\ (approx)} \\
			\hline
			5 & 1 & 128 & 256 & 1 & 128 & 128 & 0.0 & 0.9 & GeLU  
			& 255.5 (39.8) & 349.6 (54.1) & 4.17 (0.64) & 480K \\
			\hline
			&&&&&&&& 0.5 && 409.7 (30.8) & 565.0 (42.7) & 6.65 (0.50) &\\
			&&&&&&&& 0.6 && 334.5 (13.2) & 461.8 (15.9) & 5.43 (0.22) &\\
			&&&&&&&& 0.7 && 265.4 (38.0) & 371.1 (50.4) & 4.31 (0.62) &\\
			&&&&&&&& 0.8 && 233.2 (36.2) & 330.1 (49.7) & 3.78 (0.58) &\\
			&&&&&&&& 0.8 & ReLU & 439.9 (16.9) & 588.3 (20.9) & 7.18 (0.28) &\\
			\hline
			&&&&&&& 0.1 &&& 385.4 (12.8) & 513.0 (16.1) & 6.30 (0.21) &\\
			&&&&&&& 0.2 &&& 391.7 (15.6) & 521.3 (19.0) & 6.40 (0.26) &\\
			&&&&&&& 0.3 &&& 399.0 (14.9) & 530.3 (19.6) & 6.52 (0.24) &\\
			\hline
			&&&& 2 & 64 & 64 &&&& 258.6 (25.1) & 354.5 (32.9) & 4.21 (0.41) &\\
			&&&& 4 & 32 & 32 &&&& 249.7 (30.4) & 341.7 (40.2) & 4.07 (0.49) &\\
			&&&& 8 & 16 & 16 &&&& 240.4 (29.2) & 330.2 (40.3) & 3.92 (0.47) &\\
			&&&& 16 & 8 & 8 &&&& 230.1 (35.8) & 315.2 (48.0) & 3.76 (0.58) &\\
			\hline
			& 2 &&&&&&&&& 218.7 (17.3) & 302.5 (24.3) & 3.57 (0.28) & 960K\\
			& 4 &&&&&&&&& \textbf{214.4} (11.2) & \textbf{298.3} (17.1) & \textbf{3.49} (0.18) & 1.9M\\
			& 6 &&&&&&&&& 220.1 (13.4) & 307.6 (19.4) & 3.58 (0.21) & 2.8M\\
			\hline
			10 &&&&&&&&&& 268.3 (24.7) & 366.5 (32.3) & 4.37 (0.40) &\\
			20 &&&&&&&&&& 279.2 (17.2) & 378.4 (22.3) & 4.55 (0.28) &\\
			\hline
			&& 32 & 64 &&&&&&& 349.5 (42.4) & 475.1 (56.1) & 5.70 (0.69) & 31K\\
			&& 64 & 128 &&&&&&& 289.6 (23.1) & 395.9 (29.2) & 4.72 (0.38) & 121K\\
			&& 256 & 512 &&&&&&& 246.4 (15.8) & 337.2 (20.5) & 4.02 (0.26) & 1.9M\\
			&& 512 & 1024 &&&&&&& 220.8 (11.2) & 303.7 (14.9) & 3.60 (0.18) & 7.6M\\
			&& 1024 & 2048 &&&&&&& 271.6 (21.4) & 368.5 (26.6) & 4.43 (0.35) & 30M\\
			\bottomrule
		\end{tabular}
	}
	\end{table}
		
	Tables~\ref{tab:VN30-table1} and~\ref{tab:sp500-table1} summarize the forecasting performance, model hyperparameters, and parameter counts for the VN30 and S\&P 500 datasets. Several observations can be made:
	\begin{itemize}	
		\item \textbf{Effect of decay exponent ($k$):}
		Forecasting performance was highly sensitive to the decay exponent. Accuracy improved consistently as $k$ increased from 0.5 to 0.9, with the lowest errors obtained for $k=0.8$--$0.9$ on both datasets. These results demonstrate that the generalized inverse-power schedule provides an effective mechanism for controlling optimization dynamics.
		
		\item \textbf{Activation function:}
		GeLU consistently outperformed ReLU across both datasets. For example, on VN30 with $k=0.8$, GeLU reduced MAE from 37.76 to 27.60 and RMSE from 75.25 to 51.71. Similarly, on S\&P 500, MAE decreased from 439.9 to 233.2 under the same configuration. These results suggest that the smoother nonlinearity of GeLU substantially improves optimization and forecasting accuracy.
		
		\item \textbf{Dropout rate ($p_{\text{drop}}$):}
		Increasing the dropout rate beyond 0 consistently degraded forecasting performance on both datasets. The best results were obtained without dropout, indicating that additional regularization was not beneficial for this configuration.
		
		\item \textbf{Number of attention heads ($h$):}
		Increasing the number of attention heads steadily improved performance, with the best results obtained at $h=16$ for both datasets. This finding suggests that richer multi-head representations enable the model to capture complementary temporal patterns more effectively.
		
		\item \textbf{Model depth ($N$):}
		The impact of depth differed across datasets. For VN30, increasing depth beyond a single encoder--decoder layer provided little benefit and slightly degraded performance. In contrast, the S\&P 500 achieved its best results with deeper models ($N=4$), indicating that additional layers may be advantageous for more stable market dynamics.
		
		\item \textbf{Model dimension ($d_{\text{model}}$):}
		Larger model dimensions generally improved forecasting accuracy. However, the gains diminished as parameter counts increased substantially. For example, increasing $d_{\text{model}}$ from 512 to 1024 increased the parameter count from 7.6M to 30M while degrading performance, highlighting the importance of balancing accuracy and model complexity.
		
		\item \textbf{Sequence length ($L$):} A sequence length of $L=5$ consistently yielded the best forecasting performance, whereas longer contexts ($L=10,20$) did not improve accuracy.
		
		\item \textbf{Comparison with the baseline Transformer:}
		The generalized inverse-power schedule combined with GeLU activation substantially outperformed the baseline Transformer. Under comparable configurations, MAE decreased from 42.24 to 24.65 (approximately 42\% reduction) on VN30 and from 422.5 to 255.5 (approximately 40\% reduction) on the S\&P 500 dataset, demonstrating the effectiveness of the proposed architectural and optimization modifications.
	\end{itemize}
		
	\subsection{Modified Transformer with Cosine Annealing with Warmup}
	
	The modified Transformer was trained using a cosine annealing learning-rate schedule with warmup, configured with $\text{warmup\_steps}=3000$, $\text{train\_steps}=15000$, $lr_{base}=10^{-4}$, and $lr_{min}=10^{-6}$. 
	
	\begin{table}[ht]
		\centering
		\caption{Evaluation metrics for the \textbf{VN30 dataset}, averaged over 10 independent runs of the modified Transformer with the cosine annealing with warmup scheduler.}
		\label{tab:VN30-table2}
		\scriptsize
		\setlength{\tabcolsep}{3pt}
		
		\resizebox{\textwidth}{!}{
		\begin{tabular}{c c c c c c c c c | c c c | c}
			\toprule
			$L$ & $N$ & $d_{\text{model}}$ & $d_{\text{ff}}$ & $h$ & $d_k$ & $d_v$ & $p_{\text{drop}}$ & Act. Fun. &
			MAE (SD) & RMSE (SD) & MAPE (SD) & \makecell{params \\ (approx)} \\
			\hline
			5 & 1 & 128 & 256 & 1 & 128 & 128 & 0.0 & GeLU  
			& 28.79 (0.99) & 54.48 (2.35) & 1.72 (0.05) & 480K\\
			\hline
			&&&&&&& 0.1 && 31.32 (1.87) & 60.67 (4.63) & 1.86 (0.10) &\\
			&&&&&&& 0.2 && 30.62 (1.15) & 58.81 (3.19) & 1.82 (0.06) &\\
			&&&&&&& 0.3 && 30.12 (1.89) & 57.51 (4.83) & 1.80 (0.10) &\\
			&&&&&&& 0.4 && 28.67 (1.21) & 54.01 (3.22) & 1.73 (0.07) &\\
			&&&&&&& 0.5 && 27.41 (1.76) & 51.34 (3.39) & 1.66 (0.11) &\\
			&&&&&&& 0.6 && 25.15 (1.03) & 46.20 (1.79) & 1.54 (0.06) &\\
			&&&&&&& 0.7 && 22.20 (1.27) & 39.73 (2.89) & 1.38 (0.07) &\\
			&&&&&&& 0.8 && 19.91 (4.47) & 30.31 (5.70) & 1.34 (0.31) &\\
			&&&&&&& 0.9 && \textbf{17.45} (2.31) & \textbf{24.18} (2.75) & \textbf{1.23} (0.16) &\\
			&&&&&&& 0.9 & ReLU & 30.74 (4.22) & 48.01 (5.59) & 2.00 (0.30) &\\
			\hline
			&&&& 2 & 64 & 64 &&& 29.12 (0.60) & 55.26 (1.42) & 1.74 (0.03) &\\
			&&&& 4 & 32 & 32 &&& 28.59 (1.52) & 54.05 (3.54) & 1.71 (0.08) &\\
			&&&& 8 & 16 & 16 &&& 28.87 (0.70) & 54.65 (1.67) & 1.73 (0.04) &\\
			&&&& 8 & 16 & 16 & 0.9 && 18.09 (2.45) & 24.90 (3.33) & 1.28 (0.16) &\\
			\hline
			& 2 &&&&&&&& 29.19 (0.93) & 55.48 (2.24) & 1.74 (0.05) & 960K\\
			& 4 &&&&&&&& 31.76 (2.03) & 61.85 (4.96) & 1.88 (0.11) & 1.9M\\
			& 4 &&&&&& 0.9 && 23.28 (0.84) & 18.03 (1.53) & 1.44 (0.06) & 1.9M\\
			\hline
			10 &&&&&&&&& 29.18 (0.66) & 55.20 (1.57) & 1.74 (0.04) &\\
			20 &&&&&&&&& 28.90 (0.59) & 54.07 (1.39) & 1.73 (0.03) &\\
			\hline
			&& 32 & 64 &&&&&& 33.91 (2.21) & 67.16 (5.44) & 1.99 (0.12) & 31K\\
			&& 64 & 128 &&&&&& 30.10 (1.43) & 57.71 (3.36) & 1.79 (0.08) & 121K\\
			&& 256 & 512 &&&&&& 29.35 (1.88) & 55.75 (4.47) & 1.75 (0.10) & 1.9M\\
			&& 512 & 1024 &&&&&& 33.02 (2.04) & 64.52 (4.83) & 1.95 (0.11) & 7.6M\\
			\bottomrule
		\end{tabular}
	}
	\end{table}
	
	\begin{table}[ht]
		\centering
		\caption{Evaluation metrics for the \textbf{S\&P 500 dataset}, averaged over 10 independent runs of the modified Transformer with the cosine annealing with warmup scheduler.}
		\label{tab:SP500-table2}
		\scriptsize
		\setlength{\tabcolsep}{3pt}
		
		\resizebox{\textwidth}{!}{
		\begin{tabular}{c c c c c c c c c | c c c | c}
			\toprule
			$L$ & $N$ & $d_{\text{model}}$ & $d_{\text{ff}}$ & $h$ & $d_k$ & $d_v$ & $p_{\text{drop}}$ & Act. Fun. &
			MAE (SD) & RMSE (SD) & MAPE (SD) & \makecell{params \\ (approx)} \\
			\hline
			5 & 1 & 128 & 256 & 1 & 128 & 128 & 0.0 & GeLU  
			& 271.8 (27.2) & 380.2 (35.9) & 4.41 (0.44) & 480K\\
			\hline
			&&&&&&& 0.1 && 332.7 (20.0) & 461.9 (25.3) & 5.40 (0.33) &\\
			&&&&&&& 0.2 && 295.0 (26.0) & 415.4 (33.3) & 4.78 (0.42) &\\
			&&&&&&& 0.3 && 285.0 (15.8) & 404.6 (19.6) & 4.61 (0.26) &\\
			&&&&&&& 0.4 && 277.4 (20.0) & 396.7 (25.3) & 4.48 (0.33) &\\
			&&&&&&& 0.5 && 267.5 (27.4) & 384.9 (35.1) & 4.32 (0.45) &\\
			&&&&&&& 0.6 && 262.5 (22.4) & 378.4 (30.6) & 4.24 (0.36) &\\
			&&&&&&& 0.7 && 227.0 (20.6) & 335.2 (26.2) & 3.66 (0.33) &\\
			&&&&&&& 0.8 && 187.6 (24.6) & 287.4 (32.3) & 3.02 (0.40) &\\
			&&&&&&& 0.9 && \textbf{122.2} (21.5) & \textbf{189.4} (35.6) & \textbf{2.00} (0.33) &\\
			&&&&&&& 0.9 & ReLU & 181.6 (28.1) & 272.1 (41.6) & 2.96 (0.43) &\\
			\hline
			&&&& 2 & 64 & 64 &&& 279.2 (34.3) & 390.3 (44.6) & 4.53 (0.56) &\\
			&&&& 4 & 32 & 32 &&& 277.6 (35.6) & 388.8 (46.0) & 4.50 (0.58) &\\
			&&&& 8 & 16 & 16 &&& 272.4 (35.3) & 382.0 (45.9) & 4.42 (0.58) &\\
			&&&& 16 & 8 & 8 &&& 267.7 (30.7) & 376.1 (40.8) & 4.34 (0.50) &\\
			\hline
			& 2 &&&&&&&& 238.3 (44.5) & 337.4 (58.9) & 3.86 (0.72) & 960K\\
			& 4 &&&&&&&& 219.8 (56.3) & 313.3 (77.0) & 3.56 (0.91) & 1.9M\\
			& 6 &&&&&&&& 204.7 (56.6) & 292.3 (78.3) & 3.32 (0.91) & 2.8M\\
			& 6 &&&&&& 0.9 && 205.1 (6.01) & 289.0 (8.52) & 3.34 (0.10) & 2.8M\\
			\hline
			10 &&&&&&&&& 283.3 (25.9) & 393.7 (33.5) & 4.60 (0.42) &\\
			20 &&&&&&&&& 282.7 (23.5) & 391.9 (30.4) & 4.58 (0.38) &\\
			\hline
			&& 32 & 64 &&&&&& 298.1 (37.7) & 421.6 (50.2) & 4.83 (0.61) & 31K\\
			&& 64 & 128 &&&&&& 250.9 (44.0) & 356.8 (59.2) & 4.06 (0.71) & 121K\\
			&& 256 & 512 &&&&&& 301.0 (18.1) & 416.5 (23.6) & 4.89 (0.30) & 1.9M\\
			&& 512 & 1024 &&&&&& 317.6 (13.7) & 437.0 (17.4) & 5.16 (0.22) & 7.6M\\
			\bottomrule
		\end{tabular}
	}
	\end{table}
	
	Tables~\ref{tab:VN30-table2} and~\ref{tab:SP500-table2} present the forecasting results for the VN30 and S\&P 500 datasets. Several observations can be made:
	\begin{itemize}
		\item \textbf{Dropout rate ($p_{\text{drop}}$):}
		Forecasting performance improved substantially as the dropout rate increased, with the best results obtained at $p_{\text{drop}}=0.9$ on both datasets. For VN30, MAE decreased from 28.79 to 17.45 and RMSE from 54.48 to 24.18. Similarly, for S\&P 500, MAE decreased from 271.8 to 122.2 and RMSE from 380.2 to 189.4. These results indicate that strong regularization is particularly effective when combined with cosine annealing and warmup.
		
		\item \textbf{Activation function:}
		GeLU consistently outperformed ReLU under otherwise identical settings. With $p_{\text{drop}}=0.9$, GeLU reduced MAE from 30.74 to 17.45 on VN30 and from 181.6 to 122.2 on S\&P 500. This finding further supports the advantage of GeLU for Transformer-based financial forecasting.
		
		\item \textbf{Number of attention heads ($h$):}
		Increasing the number of attention heads produced modest but consistent improvements. For both datasets, performance generally improved as $h$ increased from 1 to 16, suggesting that richer multi-head representations help capture complementary temporal patterns. However, the magnitude of improvement was smaller than that obtained from optimizing dropout or activation functions.
		
		\item \textbf{Model depth ($N$):}
		The effect of depth differed across datasets. For VN30, increasing depth beyond a single encoder--decoder layer generally degraded performance under the default configuration. In contrast, deeper models consistently improved forecasting accuracy on the S\&P 500 dataset, with the best results achieved at $N=6$. This suggests that deeper architectures may be more beneficial for stable and structured market dynamics.
		
		\item \textbf{Model dimension ($d_{\text{model}}$):}
		Increasing $d_{\text{model}}$ did not consistently improve forecasting accuracy. Although larger dimensions improved performance relative to very small models ($d_{\text{model}}=32$ or $64$), the best results were generally achieved with moderate dimensions ($d_{\text{model}}=128$--$256$). Further increases to $d_{\text{model}}=512$ substantially increased parameter counts while providing limited or negative performance gains.
		
		\item \textbf{Sequence length ($L$):}
		Short contexts ($L=5$) again yielded the best results, while longer contexts ($L=10,20$) did not improve accuracy.
		
		\item \textbf{Comparison with generalized inverse-power scheduling:}
		Cosine annealing with warmup produced substantially lower forecasting errors than the generalized inverse-power scheduler. Under comparable configurations, MAE decreased from 24.65 to 17.45 (approximately 30\% reduction) on VN30 and from 255.5 to 122.2 (approximately 50\% reduction) on the S\&P 500 dataset. These results suggest that the smoother learning-rate decay provided by cosine annealing improves optimization and generalization for Transformer-based stock index forecasting.
	\end{itemize}
		
	\subsection{Modified Transformer with Cosine Annealing and SDA}
	
	The modified Transformer was trained using a cosine annealing learning-rate schedule with warmup, configured with $\text{warmup\_steps}=3000$, $\text{train\_steps}=15000$, $lr_{base}=10^{-4}$, and $lr_{min}=10^{-6}$. The proposed SDA augmentation was applied with an offset constant of $c=750$ for VN30 and $c=4000$ for S\&P 500. 
	
	\begin{table}[htbp]
		\centering
		\caption{Evaluation metrics for the \textbf{VN30 dataset}, averaged over 10 independent runs of the modified Transformer trained using cosine annealing with warmup and SDA.}
		\label{tab:VN30-table3}
		\scriptsize
		\setlength{\tabcolsep}{3pt}
		
		\resizebox{\textwidth}{!}{
		\begin{tabular}{c c c c c c c c c | c c c | c}
			\toprule
			$L$ & $N$ & $d_{\text{model}}$ & $d_{\text{ff}}$ & $h$ & $d_k$ & $d_v$ & $p_{\text{drop}}$ & Act. Fun. &
			MAE (SD) & RMSE (SD) & MAPE (SD) & \makecell{params \\ (approx)} \\
			\hline
			5 & 1 & 128 & 256 & 1 & 128 & 128 & 0.0 & GeLU  
			& 11.33 (0.0034) & 17.29 (0.0023) & 0.80 (0.0003) & 480K\\
			\hline
			&&&&&&& 0.1 && 11.29 (0.0327) & 17.26 (0.0178) & 0.80 (0.0021) &\\
			&&&&&&& 0.2 && 11.26 (0.0138) & 17.27 (0.0436) & 0.80 (0.0009) &\\
			&&&&&&& 0.3 && \textbf{11.26} (0.0347) & \textbf{17.24} (0.0156) & \textbf{0.80} (0.0024) &\\
			&&&&&&& 0.4 && 11.28 (0.0350) & 17.28 (0.0371) & 0.80 (0.0027) &\\
			&&&&&&& 0.1 & ReLU & 11.35 (0.0197) & 17.28 (0.0153) & 0.81 (0.0012) &\\
			\hline
			&&&& 2 & 64 & 64 &&& 11.33 (0.0053) & 17.29 (0.0024) & 0.80 (0.0004) &\\
			&&&& 4 & 32 & 32 &&& 11.33 (0.0021) & 17.29 (0.0010) & 0.80 (0.0002) &\\
			&&&& 8 & 16 & 16 &&& 11.33 (0.0025) & 17.29 (0.0019) & 0.80 (0.0002) &\\
			&&&& 8 & 16 & 16 & 0.1 && 11.29 (0.0197) & 17.26 (0.0227) & 0.80 (0.0013) &\\
			\hline
			& 2 &&&&&&&& 11.33 (0.0039) & 17.29 (0.0018) & 0.80 (0.0003) & 960K\\
			& 4 &&&&&&&& 11.33 (0.0041) & 17.29 (0.0029) & 0.80 (0.0004) & 1.9M\\
			\hline
			10 &&&&&&&&& 11.37 (0.0025) & 17.34 (0.0010) & 0.81 (0.0002) &\\
			20 &&&&&&&&& 11.46 (0.0026) & 17.46 (0.0011) & 0.81 (0.0002) &\\
			\hline
			&& 32 & 64 &&&&&& 11.33 (0.0181) & 17.30 (0.0131) & 0.80 (0.0015) & 31K\\
			&& 64 & 128 &&&&&& 11.33 (0.0033) & 17.29 (0.0020) & 0.80 (0.0003) & 121K\\
			&& 256 & 512 &&&&&& 11.34 (0.0031) & 17.29 (0.0026) & 0.80 (0.0002) & 1.9M\\
			&& 512 & 1024 &&&&&& 11.34 (0.0021) & 17.29 (0.0031) & 0.80 (0.0002) & 7.6M\\
			\bottomrule
		\end{tabular}
	}
	\end{table}
	
	\begin{table}[htbp]
		\centering
		\caption{Evaluation metrics for the \textbf{S\&P 500 dataset}, averaged over 10 independent runs of the modified Transformer trained using cosine annealing with warmup and SDA.}
		\label{tab:SP500-table3}
		\scriptsize
		\setlength{\tabcolsep}{3pt}
		
		\resizebox{\textwidth}{!}{
		\begin{tabular}{c c c c c c c c c | c c c | c}
			\toprule
			$L$ & $N$ & $d_{\text{model}}$ & $d_{\text{ff}}$ & $h$ & $d_k$ & $d_v$ & $p_{\text{drop}}$ & Act. Fun. &
			MAE (SD) & RMSE (SD) & MAPE (SD) & \makecell{params \\ (approx)} \\
			\hline
			5 & 1 & 128 & 256 & 1 & 128 & 128 & 0.0 & GeLU  
			& 35.31 (0.0032) & 52.09 (0.0124) & 0.64 (0.0001) & 480K\\
			\hline
			&&&&&&& 0.1 && 35.22 (0.0414) & 52.11 (0.1039) & 0.64 (0.0007) &\\
			&&&&&&& 0.2 && 35.21 (0.2530) & 52.24 (0.4309) & 0.64 (0.0040) &\\
			&&&&&&& 0.3 && 35.21 (0.2086) & 52.31 (0.3688) & 0.64 (0.0038) &\\
			&&&&&&& 0.4 && 35.31 (0.4362) & 52.43 (0.6729) & 0.64 (0.0075) &\\
			&&&&&&& 0.1 & ReLU & 35.27 (0.0742) & 52.04 (0.0977) & 0.64 (0.0013) &\\
			\hline
			&&&& 2 & 64 & 64 &&& 35.31 (0.0024) & 52.09 (0.0102) & 0.64 (0.0000) &\\
			&&&& 4 & 32 & 32 &&& 35.31 (0.0033) & 52.08 (0.0094) & 0.64 (0.0001) &\\
			&&&& 8 & 16 & 16 &&& 35.31 (0.0046) & 52.09 (0.0128) & 0.64 (0.0001) &\\
			&&&& 8 & 16 & 16 & 0.1 && \textbf{35.21} (0.0446) & \textbf{52.06} (0.0402) & \textbf{0.64} (0.0008) &\\
			\hline
			& 2 &&&&&&&& 35.32 (0.0059) & 52.09 (0.0111) & 0.64 (0.0001) & 960K\\
			& 4 &&&&&&&& 35.30 (0.0074) & 52.07 (0.0066) & 0.64 (0.0002) & 1.9M\\
			\hline
			10 &&&&&&&&& 35.41 (0.0049) & 52.24 (0.0135) & 0.64 (0.0001) &\\
			20 &&&&&&&&& 35.62 (0.0060) & 52.55 (0.0169) & 0.65 (0.0001) &\\
			\hline
			&& 32 & 64 &&&&&& 35.32 (0.0301) & 52.09 (0.0355) & 0.64 (0.0006) & 31K\\
			&& 64 & 128 &&&&&& 35.32 (0.0171) & 52.08 (0.0202) & 0.64 (0.0003) & 121K\\
			&& 256 & 512 &&&&&& 35.32 (0.0063) & 52.10 (0.0119) & 0.64 (0.0001) & 1.9M\\
			&& 512 & 1024 &&&&&& 35.33 (0.0066) & 52.11 (0.0097) & 0.64 (0.0001) & 7.6M\\
			\bottomrule
		\end{tabular}
	}
	\end{table}
	
	Tables~\ref{tab:VN30-table3} and~\ref{tab:SP500-table3} present the forecasting results for both datasets. Several observations can be made:
	\begin{itemize}
		\item \textbf{Impact of SDA:}
		SDA produced substantial improvements over all previously evaluated configurations. Relative to the best-performing model without SDA, MAE decreased from 17.45 to 11.26 (approximately 35\% reduction) on VN30 and from 122.2 to 35.21 (approximately 71\% reduction) on the S\&P 500 dataset. These results demonstrate that SDA substantially enhances model generalization and is responsible for the largest performance improvements observed in this study.
		
		\item \textbf{Training stability:}
		In addition to improving forecasting accuracy, SDA substantially reduced run-to-run variability. Across both datasets, the standard deviations of MAE, RMSE, and MAPE decreased by several orders of magnitude compared with the non-SDA configurations. For example, on VN30 the standard deviation of MAE decreased from 2.31 under cosine annealing without SDA to 0.03, while on the S\&P 500 dataset it decreased from 21.5 to 0.04. These results indicate that SDA produces highly stable and reproducible training outcomes.
		
		\item \textbf{Reduced sensitivity to hyperparameters:}
		Unlike the previous configurations, forecasting performance remained remarkably stable across a wide range of hyperparameter settings. Variations in dropout rate, number of attention heads, model depth, and projection dimension resulted in only negligible changes in MAE, RMSE, and MAPE. This indicates that SDA substantially improves training robustness and reduces dependence on careful hyperparameter tuning.
		
		\item \textbf{Dropout and activation functions:}
		The influence of dropout became minimal after applying SDA. Similarly, GeLU and ReLU produced nearly identical results, with only marginal differences across evaluation metrics. These findings suggest that SDA itself provides a strong regularizing effect, reducing sensitivity to both regularization strength and activation choice.
		
		\item \textbf{Attention heads:}
		Performance remained virtually unchanged as the number of attention heads increased from 1 to 8. In contrast to earlier experiments, larger multi-head configurations no longer provided measurable benefits, indicating that SDA simplifies the forecasting task and reduces the need for complex attention structures.
		
		\item \textbf{Model depth and dimension:}
		Increasing model depth or projection dimension yielded little improvement despite substantially increasing parameter counts. Even compact models with only 31K parameters achieved performance comparable to models containing several million parameters. This highlights the parameter efficiency enabled by SDA and suggests that accurate forecasting can be achieved without large Transformer architectures.
		
		\item \textbf{Sequence length:}
		The shortest input window ($L=5$) again achieved the best results, while longer contexts produced slightly higher errors. This finding is consistent with previous experiments and suggests that predictive information is concentrated in recent observations.
	\end{itemize}
	
	\begin{table}[htbp]
		\centering
		\caption{Forecasting accuracy and run-to-run variability across 10 independent runs. 
			Results correspond to the best-performing hyperparameter configurations identified in Section~\ref{sec:evaluation}.}
		\label{tab:accuracy_stability}
		\scriptsize
		\setlength{\tabcolsep}{3pt}
		
		\resizebox{\textwidth}{!}{
		\begin{tabular}{l|c c c|c c c}
			\toprule
			\multirow{2}{*}{Model} & \multicolumn{3}{c|}{VN30} & \multicolumn{3}{c}{S\&P 500} \\
			& MAE (SD) & RMSE (SD) & MAPE (SD) & MAE (SD) & RMSE (SD) & MAPE (SD) \\
			\hline
			Baseline Transformer
			& 42.24 (3.74) & 78.70 (6.33) & 2.55 (0.26)
			& 422.5 (41.1) & 600.9 (56.2) & 6.87 (0.65) \\
			
			Modified Transformer + Generalized Inverse-Power
			& 22.13 (1.66) & 39.24 (3.74) & 1.37 (0.09)
			& 214.4 (11.2) & 298.3 (17.1) & 3.49 (0.18) \\
			
			Modified Transformer + Cosine Annealing
			& 17.45 (2.31) & 24.18 (2.75) & 1.23 (0.16)
			& 122.2 (21.1) & 189.4 (35.6) & 2.00 (0.33) \\
			
			Modified Transformer + Cosine Annealing + SDA
			& 11.26 (0.03) & 17.24 (0.02) & 0.80 (0.002)
			& 35.21 (0.04) & 52.06 (0.04) & 0.64 (0.0008) \\
			\bottomrule
		\end{tabular}
	}
	\end{table}
	
	\begin{figure}[ht]
		\centering
		\includegraphics[width=\linewidth]{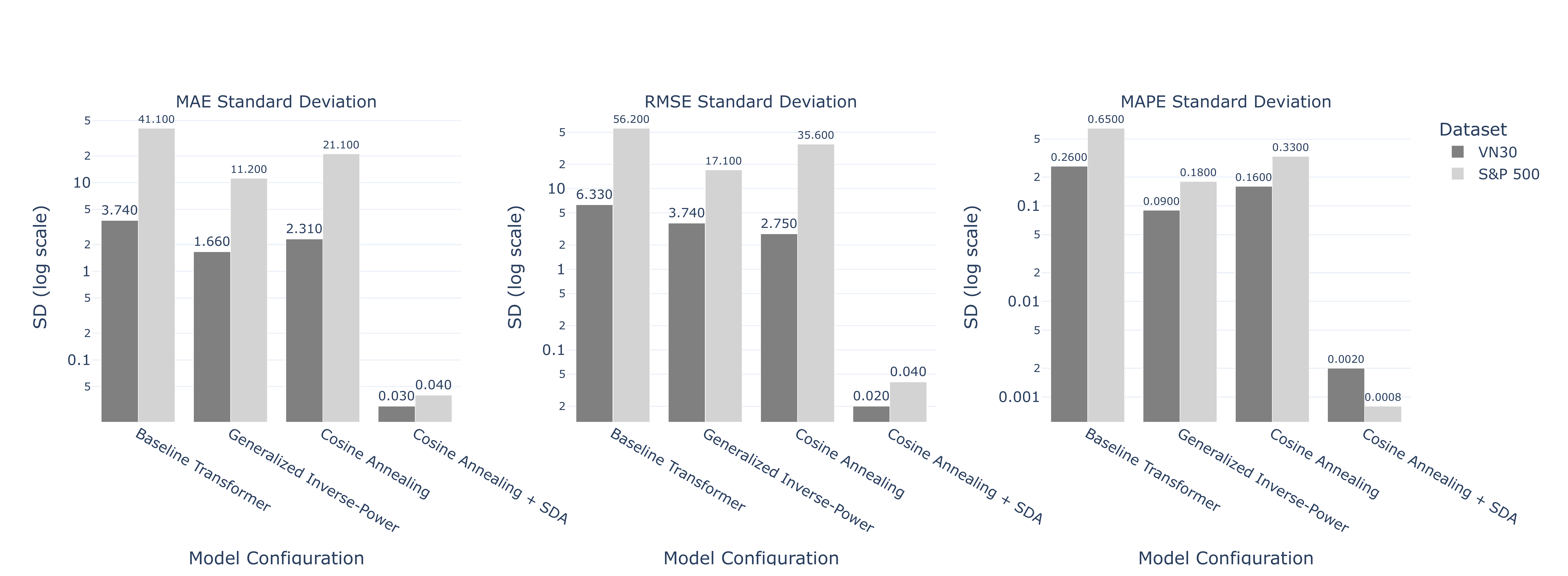}
		\caption{Run-to-run variability across 10 independent runs measured by the standard deviations of MAE, RMSE, and MAPE for VN30 and S\&P 500. All subplots use logarithmic scaling on the vertical axis.}
		\label{fig:forecasting_variability}
	\end{figure}
	
	Table~\ref{tab:accuracy_stability} and Figure~\ref{fig:forecasting_variability} summarize the effects of the proposed modifications on forecasting accuracy and run-to-run stability across 10 independent experiments. While both the generalized inverse-power scheduler and cosine annealing substantially reduced forecasting errors relative to the baseline Transformer, the incorporation of SDA produced the largest improvements in both predictive accuracy and reproducibility. Across all evaluation metrics (MAE, RMSE, and MAPE), the SDA-enhanced configuration achieved dramatically lower standard deviations than competing approaches. 
	In particular, the standard deviation of MAE decreased from 3.74 to 0.03 on VN30 and from 41.1 to 0.04 on the S\&P 500. Similar reductions were observed for RMSE and MAPE, as illustrated in Figure~\ref{fig:forecasting_variability}. The logarithmic-scale plots further highlight the substantial reduction in forecasting variability achieved by SDA across both datasets.
	Notably, the dramatic reduction in variability was accompanied by substantial improvements in forecasting accuracy, suggesting that SDA not only enhances generalization but also makes Transformer training considerably less sensitive to random initialization and optimization dynamics. These results indicate that the proposed augmentation strategy improves both the effectiveness and reliability of Transformer-based financial forecasting.

	%=================================================================
	\section{Benchmark Analysis}\label{sec:results}
	
	The benchmark results reported in this section correspond to representative single-run experiments using the best-performing hyperparameter configurations identified in Section~\ref{sec:evaluation}. While the following figures provide a visual comparison of forecast trajectories, the robustness of these results has already been established through the multi-run evaluation. In particular, the SDA-enhanced models exhibited extremely low run-to-run variability (Table~\ref{tab:accuracy_stability}), indicating that the benchmark forecasts shown here are highly representative rather than isolated outcomes.
	
	\begin{table}[htbp]
		\centering
		\caption{Comparison of the best-performing Transformer configurations on the VN30 and S\&P 500 datasets. Hyperparameters for each model were selected according to the results reported in Section~\ref{sec:evaluation}.}
		\label{tab:summary}
		\scriptsize
		\setlength{\tabcolsep}{3pt}
		
		\resizebox{\textwidth}{!}{
		\begin{tabular}{l|c c c|c c c}
			\toprule
			& \multicolumn{3}{c|}{VN30} & \multicolumn{3}{c}{S\&P 500} \\
			Model & MAE & RMSE & MAPE & MAE & RMSE & MAPE \\
			\hline
			Baseline Transformer
			& 39.46 & 74.23 & 2.38
			& 363.9 & 532.5 & 5.86 \\
			
			Modified Transformer + Generalized Inverse-Power
			& 20.85 & 36.32 & 1.30
			& 187.0 & 260.2 & 3.05 \\
			
			Modified Transformer + Cosine Annealing
			& 14.92 & 21.79 & 1.03
			& 109.8 & 170.5 & 1.81 \\
			
			Modified Transformer + Cosine Annealing + SDA
			& \textbf{11.28} & \textbf{17.28} & \textbf{0.80}
			& \textbf{35.20} & \textbf{52.03} & \textbf{0.64} \\
			\bottomrule
		\end{tabular}
	}
	\end{table}
	
	Table~\ref{tab:summary} reveals a consistent improvement trend across all evaluation metrics. Relative to the baseline Transformer, the combination of cosine annealing scheduling and SDA reduced MAE by 71.4\% on VN30 (39.46 $\rightarrow$ 11.28) and by 90.3\% on the S\&P 500 (363.9 $\rightarrow$ 35.20). Similar reductions were observed for RMSE and MAPE, demonstrating the complementary benefits of improved optimization and data augmentation.
	
	\begin{figure}[htbp]
		\centering
		\includegraphics[width=\linewidth]{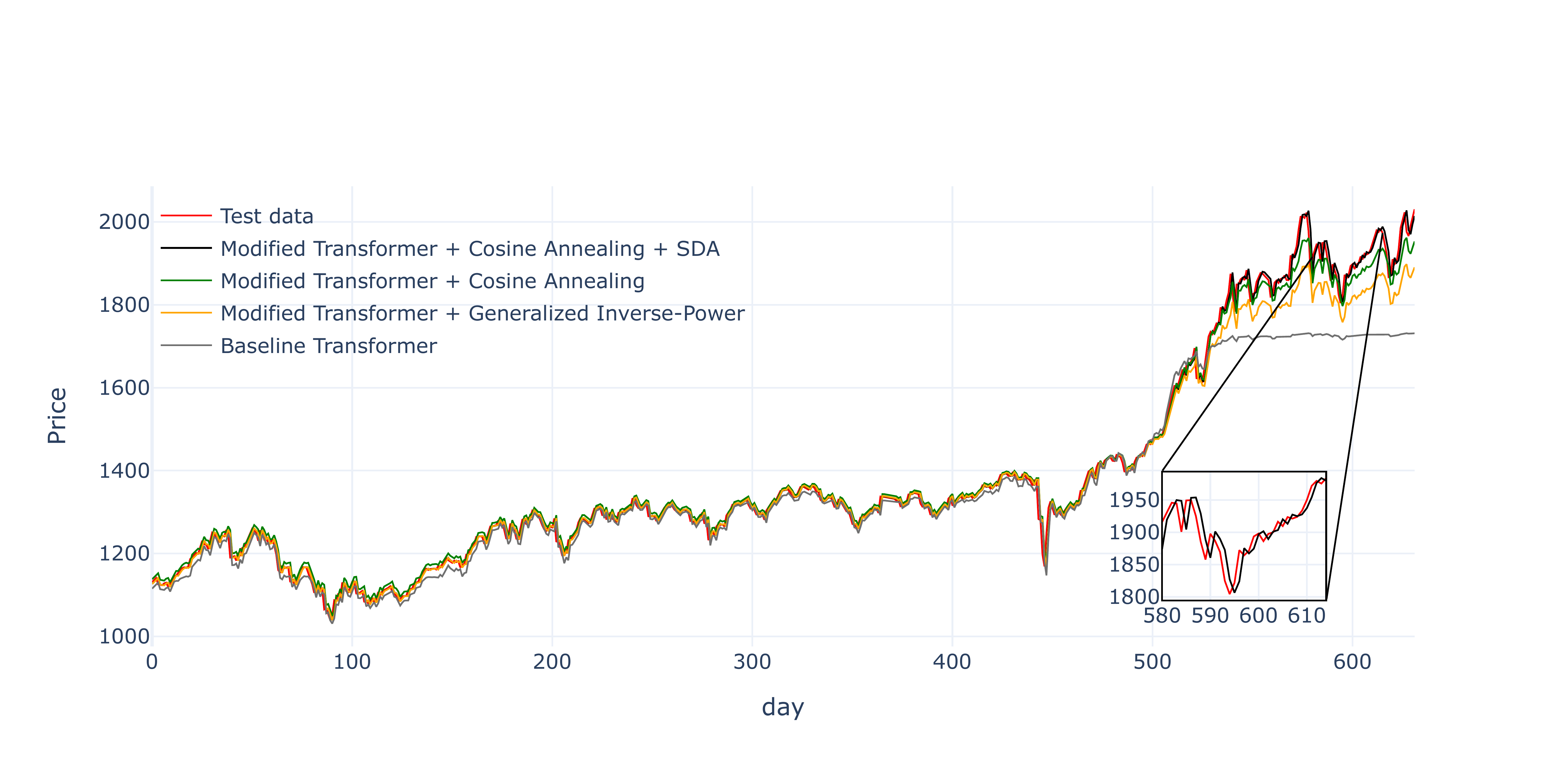}
		\caption{Comparison of VN30 test-set forecasts generated by the baseline Transformer and modified Transformer variants.} 
		\label{fig:vn30}
	\end{figure}
	
	\begin{figure}[htbp]
		\centering
		\includegraphics[width=\linewidth]{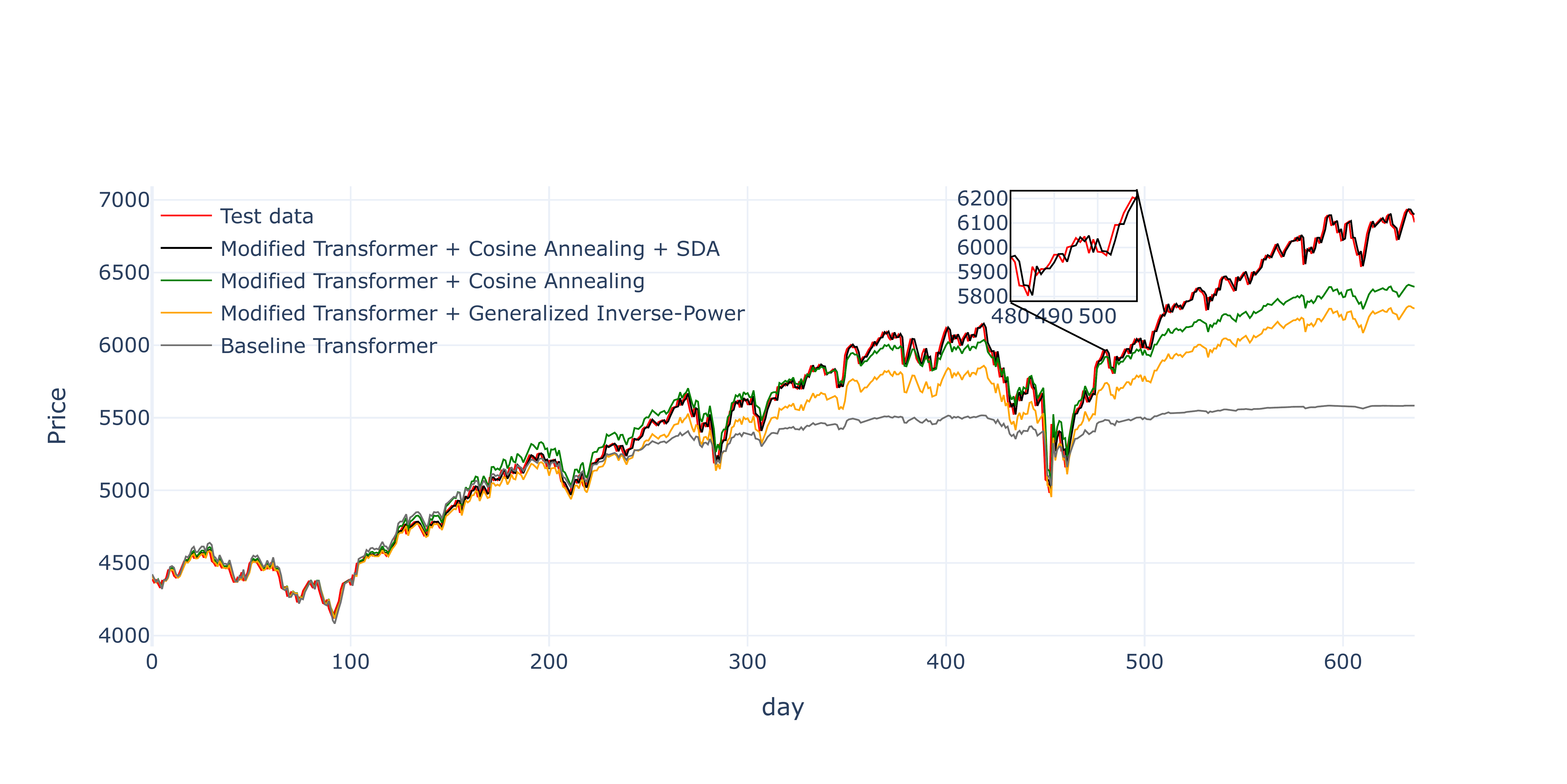}
		\caption{Comparison of S\&P 500 test-set forecasts generated by the baseline Transformer and modified Transformer variants.} 
		\label{fig:sp500}
	\end{figure}
	
	For the VN30 dataset, Figure~\ref{fig:vn30} shows that each successive modification progressively improves forecasting accuracy. The SDA-enhanced model produces predictions that most closely follow the actual test series, particularly in regions where index values extend beyond the range observed during training. This visual improvement is consistent with the substantial error reductions reported in Table~\ref{tab:summary}.
	
	For the S\&P 500 dataset, Figure~\ref{fig:sp500} reveals an even larger improvement. The modified Transformer with cosine annealing scheduling and SDA closely tracks the actual index trajectory and remains accurate when index values move beyond the historical training range, whereas the baseline Transformer exhibits substantially larger forecasting errors. Together with the stability results in Table~\ref{tab:accuracy_stability}, these findings indicate that the performance gains are both accurate and highly reproducible across independent training runs.
	
	%=================================================================
	\section{Conclusions and Future Work}\label{sec:concl}
	
	This paper presented a modified Transformer framework for one-step stock market index forecasting, combining architectural refinements, advanced learning-rate scheduling strategies, and the proposed Shifted Data Augmentation (SDA) technique. The framework was evaluated on two benchmark datasets representing both an emerging market (VN30) and a mature market (S\&P 500). Experimental results demonstrated that the proposed modifications consistently improved forecasting performance relative to the baseline Transformer. In particular, cosine annealing with warmup achieved lower forecasting errors than the generalized inverse-power scheduler across both datasets, indicating the importance of effective optimization strategies in Transformer-based financial forecasting.
	
	The most significant contribution of this study is the proposed SDA technique. By augmenting training sequences through constant-offset transformations that preserve temporal structure, SDA substantially improved forecasting accuracy while dramatically reducing run-to-run variability. The combination of cosine annealing and SDA achieved the best performance on both datasets and exhibited strong robustness when test values extended beyond the range observed during training. These results suggest that exposing the model to a broader range of value levels during training can be more beneficial than increasing model complexity alone.
	
	Beyond the primary forecasting results, two important empirical findings emerged. First, shorter input sequences consistently outperformed longer contexts, suggesting that stock index forecasting is characterized by strong short-memory dynamics. Second, the reduced hyperparameter sensitivity observed under SDA indicates that the proposed augmentation strategy improves not only predictive performance but also training robustness, allowing relatively compact Transformer configurations to achieve competitive results across diverse settings.
	
	Several directions remain open for future work. First, although this study focused on the Transformer framework, future research should investigate whether the benefits of SDA generalize to other forecasting architectures, including RNN, LSTM, GRU, and CNN models. Such investigations would require careful architecture-specific hyperparameter optimization and provide a broader assessment of the general applicability of SDA. Second, SDA could be integrated with more recent Transformer variants developed for time-series forecasting to evaluate its effectiveness under alternative attention mechanisms and architectural designs. Finally, future work may extend the proposed framework to multivariate, multi-step, and probabilistic forecasting settings incorporating additional information sources such as technical indicators, macroeconomic variables, and sentiment-based features.
	%=================================================================
	\section*{Abbreviations}
	
	The abbreviations used throughout this manuscript are summarized in Table~\ref{tab:abbreviations}.
	
	\begin{table}[htbp]
		\centering
		\caption{List of abbreviations used in this manuscript.}
		\label{tab:abbreviations}
		\small
		\begin{tabular}{ll}
			\toprule
			\textbf{Abbreviation} & \textbf{Definition} \\
			\hline
			MAE   & Mean Absolute Error \\
			MSE   & Mean Squared Error \\
			RMSE  & Root Mean Squared Error \\
			MAPE  & Mean Absolute Percentage Error \\
			VN30  & Index of the 30 Largest and Most Liquid Stocks on the Ho Chi Minh Stock Exchange \\
			S\&P 500 & Standard \& Poor's 500 Index \\
			SDA   & Shifted Data Augmentation \\
			ReLU  & Rectified Linear Unit \\
			GeLU  & Gaussian Error Linear Unit \\
			NLP   & Natural Language Processing \\
			RNN   & Recurrent Neural Network \\
			LSTM  & Long Short-Term Memory \\
			GRU   & Gated Recurrent Unit \\
			FFN   & Feed-Forward Network \\
			CNN   & Convolutional Neural Network\\
			\bottomrule
		\end{tabular}
	\end{table}
	
	%=================================================================
	\section*{Data Availability}
	
	The datasets analyzed in this study are publicly available from Investing.com. 
	Historical data for the VN30 and S\&P 500 indices can be obtained from:
	
	\begin{itemize}
		\item \href{https://vn.investing.com/indices/vn-30-historical-data}{Investing.com: VN30 Historical Index Data}
		\item \href{https://vn.investing.com/indices/us-spx-500-historical-data}{Investing.com: S\&P 500 Historical Index Data}
	\end{itemize}

	%=================================================================
	

\begin{thebibliography}{99}
		
		\bibitem{sezer2019survey}
		O.~B. Sezer, M.~U. Gudelek, and A.~M. Ozbayoglu,
		``Financial time series forecasting with deep learning: A systematic literature review: 2005--2019,''
		\emph{Applied Soft Computing}, vol.~90, p.~106181, 2020.
		
		\bibitem{buczynski2023review}
		M.~Buczyński, M.~Chlebus, K.~Kopczewska, and M.~Zajenkowski,
		``Financial time series models---Comprehensive review of deep learning approaches and practical recommendations,''
		\emph{Engineering Proceedings}, vol.~39, no.~1, p.~79, 2023.
		
		\bibitem{kasse2025enhancing}
		I.~Kasse, J.~Kusuma, A.~Lawi, and A.~Rahim,
		``Enhancing stock price forecasting accuracy through compositional learning of recurrent architectures: A multi-variant {RNN} approach,''
		\emph{{IEEE} Access}, vol.~13, pp.~1420--1435, 2025.
		
		\bibitem{hartanto2026attention}
		R.~D. Hartanto, G.~F. Shidik, F.~Alzami, A.~Z. Fanani, A.~Marjuni, and A.~Syukur,
		``Attention-augmented {GRU} for stock forecasting: A trade-off between directional accuracy and price prediction error,''
		\emph{Journal of Computing Theories and Applications}, vol.~3, no.~4, pp.~457--472, 2026.
		
		\bibitem{akkas2026attention}
		H.~Akkas, B.~Kolukisa, and B.~Bakir-Gungor,
		``An attention-based autoencoder model with gated recurrent unit for stock price movement prediction,''
		\emph{International Journal of Computational Intelligence Systems}, vol.~19, no.~1, p.~45, 2026.
		
		\bibitem{vaswani2017attention}
		A.~Vaswani, N.~Shazeer, N.~Parmar, J.~Uszkoreit, L.~Jones, A.~N. Gomez, \L{}.~Kaiser, and I.~Polosukhin,
		``Attention is all you need,''
		in \emph{Advances in Neural Information Processing Systems (NeurIPS)}, 2017, pp.~5998--6008.
		
		\bibitem{dosovitskiy2021image}
		A.~Dosovitskiy, L.~Beyer, A.~Kolesnikov, D.~Weissenborn, X.~Zhai, T.~Unterthiner, M.~Dehghani, M.~Minderer, G.~Heigold, S.~Gelly, J.~Uszkoreit, and N.~Houlsby, ``An image is worth 16$\times$16 words: Transformers for image recognition at scale,'' in \emph{Proceedings of the 9th International Conference on Learning Representations (ICLR)}, 2021.
		
		\bibitem{dong2018speech}
		L.~Dong, S.~Xu, and B.~Xu,
		``Speech-transformer: A no-recurrence sequence-to-sequence model for speech recognition,''
		in \emph{Proc. IEEE ICASSP}, 2018, pp.~5884--5888.
		
		\bibitem{wen2023survey}
		Q.~Wen, T.~Zhou, C.~Zhang, W.~Chen, Z.~Ma, J.~Yan, and L.~Sun,
		``Transformers in time series: A survey,''
		in \emph{Proceedings of the Thirty-Second International Joint Conference on Artificial Intelligence (IJCAI)}, pp.~6778--6786, 2023.
		
		\bibitem{ieee2023lstmvstransformer}
		S.~Venkatesan, K.~Devi, T.~V. Ambuli, and V.~Ramu,
		``Deep learning for stock market prediction: A comparative study of {LSTM} and {Transformer}-based models,''
		in \emph{Proc. IEEE International Conference on Computing}, 2023, pp.~1--6.
		
		\bibitem{acm2024transformer}
		L.~Mozaffari and J.~Zhang,
		``Predictive modeling of stock prices using {Transformer} model,''
		in \emph{Proc. International Conference on Machine Learning Technologies (ICMLT)}, 2024, pp.~41--48.
		
		\bibitem{bentaieb2012multistep}
		S.~Ben~Taieb, G.~Bontempi, A.~F. Atiya, and A.~Sorjamaa,
		``A review and comparison of strategies for multi-step ahead time series forecasting based on the {NN5} forecasting competition,''
		\emph{Expert Systems with Applications}, vol.~39, no.~8, pp.~7067--7083, 2012.
		
		\bibitem{chevillon2007direct}
		G.~Chevillon,
		``Direct multi-step estimation and forecasting,''
		\emph{Journal of Economic Surveys}, vol.~21, no.~4, pp.~746--785, 2007.
		
		\bibitem{hendrycks2016gelu}
		D.~Hendrycks and K.~Gimpel, ``Gaussian error linear units ({GELUs}),'' \emph{arXiv preprint arXiv:1606.08415}, 2016.
		
		\bibitem{hamilton1989}
		J.~D. Hamilton,
		``A new approach to the economic analysis of nonstationary time series and the business cycle,''
		\emph{Econometrica}, vol.~57, no.~2, pp.~357--384, 1989.
		
		\bibitem{kim1999}
		C.~J. Kim and C.~R. Nelson,
		\emph{State-Space Models with Regime Switching}.
		MIT Press, 1999.
		
		\bibitem{diebold1996}
		F.~X. Diebold and G.~D. Rudebusch,
		``Measuring business cycles: A modern perspective,''
		\emph{Review of Economics and Statistics}, vol.~78, no.~1, pp.~67--77, 1996.
		
		\bibitem{stock2016}
		J.~H. Stock and M.~W. Watson,
		``Dynamic factor models, factor-augmented vector autoregressions, and structural vector autoregressions in macroeconomics,''
		in \emph{Handbook of Macroeconomics}, vol.~2.
		Elsevier, 2016, pp.~415--525.
		
		\bibitem{akbal2024}
		O.~Akbal,
		``Regime-switching factor models and nowcasting with big data,''
		\emph{{IMF} Working Paper}, no. {WP}/24/42, 2024.
		
		\bibitem{lu2025}
		E.~D. Lu, C.~Findling, M.~Clausel, A.~Leite, W.~Gong, and P.~Kersaudy, ``Adaptive regime-switching forecasts with distribution-free uncertainty: Deep switching state-space models meet conformal prediction,'' \emph{arXiv preprint arXiv:2512.03298}, 2025.
		
		\bibitem{hochreiter1997long}
		S.~Hochreiter and J.~Schmidhuber,
		``Long short-term memory,''
		\emph{Neural Computation}, vol.~9, no.~8, pp.~1735--1780, 1997.
		
		\bibitem{siami2018forecasting}
		S.~Siami-Namini and A.~S. Namin, ``Forecasting economics and financial time series: ARIMA vs. LSTM,'' in \emph{Proceedings of the 2018 IEEE International Conference on Big Data (Big Data)}, 2018, pp. 5174--5182.
		
		\bibitem{siami2019comparative}
		S.~Siami-Namini, N.~Tavakoli, and A.~S. Namin, ``A comparative analysis of forecasting financial time series using ARIMA, LSTM, and BiLSTM,'' in \emph{Proceedings of the 2019 IEEE International Conference on Big Data (Big Data)}, 2019, pp. 4179--4186.
		
		\bibitem{cho2014learning}
		K.~Cho, B.~van Merri{\"e}nboer, C.~Gulcehre, D.~Bahdanau, F.~Bougares, H.~Schwenk, and Y.~Bengio, ``Learning phrase representations using {RNN} encoder--decoder for statistical machine translation,'' in \emph{Proceedings of the 2014 Conference on Empirical Methods in Natural Language Processing (EMNLP)}, 2014, pp. 1724--1734.
		
		\bibitem{fischer2018deep}
		T.~Fischer and C.~Krauss,
		``Deep learning with long short-term memory networks for financial market predictions,''
		\emph{European Journal of Operational Research}, vol.~270, no.~2, pp.~654--669, 2018.
		
		\bibitem{lu2020cnn}
		W.~Lu, J.~Li, Y.~Li, A.~Sun, and J.~Wang,
		``A {CNN}-{LSTM}-based model to forecast stock prices,''
		\emph{Complexity}, vol.~2020, p.~6622927, 2020.
		
		\bibitem{thach2025forecasting}
		T.~T. Thach,
		``Forecasting stock market indices using integration of encoder, decoder, and attention mechanism,''
		\emph{Entropy}, vol.~27, no.~1, p.~82, 2025.
		
		\bibitem{wang2022stock}
		C.~Wang, Y.~Chen, S.~Zhang, and Q.~Zhang,
		``Stock market index prediction using deep {Transformer} model,''
		\emph{Expert Systems with Applications}, vol.~208, p.~118128, 2022.
		
		\bibitem{bui2025time2vec}
		N.~K. Bui, N.~D. Chien, P.~Kov{\'a}cs, and G.~Bogn{\'a}r,
		``Transformer encoder and multi-features {Time2Vec} for financial prediction,''
		in \emph{Proc. European Signal Processing Conference (EUSIPCO)}, 2025, pp.~1682--1686.
		
		\bibitem{springer2024dualattention}
		A.~Hadizadeh, M.~J. Tarokh, and M.~M. Ghazani,
		``A novel {Transformer}-based dual attention architecture for the prediction of financial time series,''
		\emph{Journal of King Saud University -- Computer and Information Sciences}, vol.~37, no.~5, p.~102140, 2025.
		
		\bibitem{psr2023phien}
		N.~N.~Phien and J.~Platos,
		``The PSR-transformer nexus: A deep dive into stock time series forecasting,''
		\emph{International Journal of Advanced Computer Science and Applications},
		vol.~14, no.~12, 2023.
		
		
		\bibitem{gezici2024}
		A.~H.~Gezici and E.~Sefer,
		``Deep transformer-based asset price and direction prediction,''
		\emph{IEEE Access},
		vol.~12, pp.~24164--24178, 2024.
		
		\bibitem{alridhawi2026}
		M.~Al Ridhawi, M.~H.~Ali, and H.~Al Osman,
		``Stock market prediction using node transformer architecture integrated with BERT sentiment analysis,''
		\emph{IEEE Access}, 2026.
		
		\bibitem{wang2023}
		S.~Wang,
		``A stock price prediction method based on BiLSTM and improved transformer,''
		\emph{IEEE Access},
		vol.~11, pp.~104211--104223, 2023.
		
		\bibitem{friday2025}
		I.~K.~Friday, S.~P.~Pati, and D.~Mishra,
		``A multi-modal approach using a hybrid vision transformer and temporal fusion transformer model for stock price movement classification,''
		\emph{IEEE Access}, 2025.
		
		\bibitem{krizhevsky2012imagenet}
		A.~Krizhevsky, I.~Sutskever, and G.~E. Hinton,
		``{ImageNet} classification with deep convolutional neural networks,''
		in \emph{Advances in Neural Information Processing Systems (NeurIPS)}, 2012.
		
		
		\bibitem{perez2017augmentation}
		L.~Perez and J.~Wang, ``The effectiveness of data augmentation in image classification using deep learning,'' \emph{arXiv preprint arXiv:1712.04621}, 2017.
		
		\bibitem{shorten2019augmentation}
		C.~Shorten and T.~M. Khoshgoftaar,
		``A survey on image data augmentation for deep learning,''
		\emph{Journal of Big Data}, vol.~6, no.~1, pp.~1--48, 2019.
		
		\bibitem{shorten2021augmentation}
		C.~Shorten, T.~M. Khoshgoftaar, and B.~Furht,
		``Text data augmentation for deep learning,''
		\emph{Journal of Big Data}, vol.~8, no.~1, p.~101, 2021.
		
		\bibitem{zhou2024augmentation}
		Y.~Zhou, C.~Guo, X.~Wang, Y.~Chang, and Y.~Wu, ``A survey on data augmentation in the large model era,'' \emph{arXiv preprint arXiv:2401.15422}, 2024.
		
		\bibitem{iglesias2023survey}
		G.~Iglesias, E.~Talavera, \'{A}.~Gonz\'{a}lez-Prieto, A.~Mozo, and S.~G\'{o}mez-Canaval, ``Data augmentation techniques in time series domain: A survey and taxonomy,'' \emph{Neural Computing and Applications}, vol.~35, no.~13, pp. 10123--10145, 2023.
		
		\bibitem{devlin2018bert}
		J.~Devlin, M.-W. Chang, K.~Lee, and K.~Toutanova, ``{BERT}: Pre-training of deep bidirectional {Transformers} for language understanding,''
		in \emph{Proc. NAACL-HLT}, 2019, pp.~4171--4186.
		
		\bibitem{kingma2014adam}
		D.~P. Kingma and J.~Ba, ``Adam: A method for stochastic optimization,'' in \emph{Proceedings of the 3rd International Conference on Learning Representations (ICLR)}, 2015.
		
		\bibitem{loshchilov2016sgdr}
		I.~Loshchilov and F.~Hutter, ``SGDR: Stochastic gradient descent with warm restarts,'' in \emph{Proceedings of the 5th International Conference on Learning Representations (ICLR)}, 2017.
		
	\end{thebibliography}
\end{document}